\documentclass{article}
\usepackage{graphicx} % Required for inserting images
\usepackage{url}
\usepackage{xcolor}
\usepackage[numbers,sort&compress]{natbib}
\usepackage{amsmath}
\usepackage{amssymb}
\usepackage{booktabs}
\usepackage{longtable}
\usepackage{array}
\newcolumntype{L}[1]{>{\raggedright\arraybackslash}p{#1}}
\usepackage{pdflscape}
\usepackage{colortbl}
\usepackage{makecell}
\usepackage{multirow}
\usepackage[margin=1in]{geometry}
\usepackage{tcolorbox}
\usepackage{subcaption}
\usepackage{tikz}
\tcbuselibrary{skins,breakable}
\usepackage{tabularx}

\newcommand{\model}{MKB}

\title{%
  \begin{minipage}{\textwidth}
    \raggedright
    \vspace{-8em}  % 整体标题区域向上移动
    \includegraphics[width=0.3\textwidth]{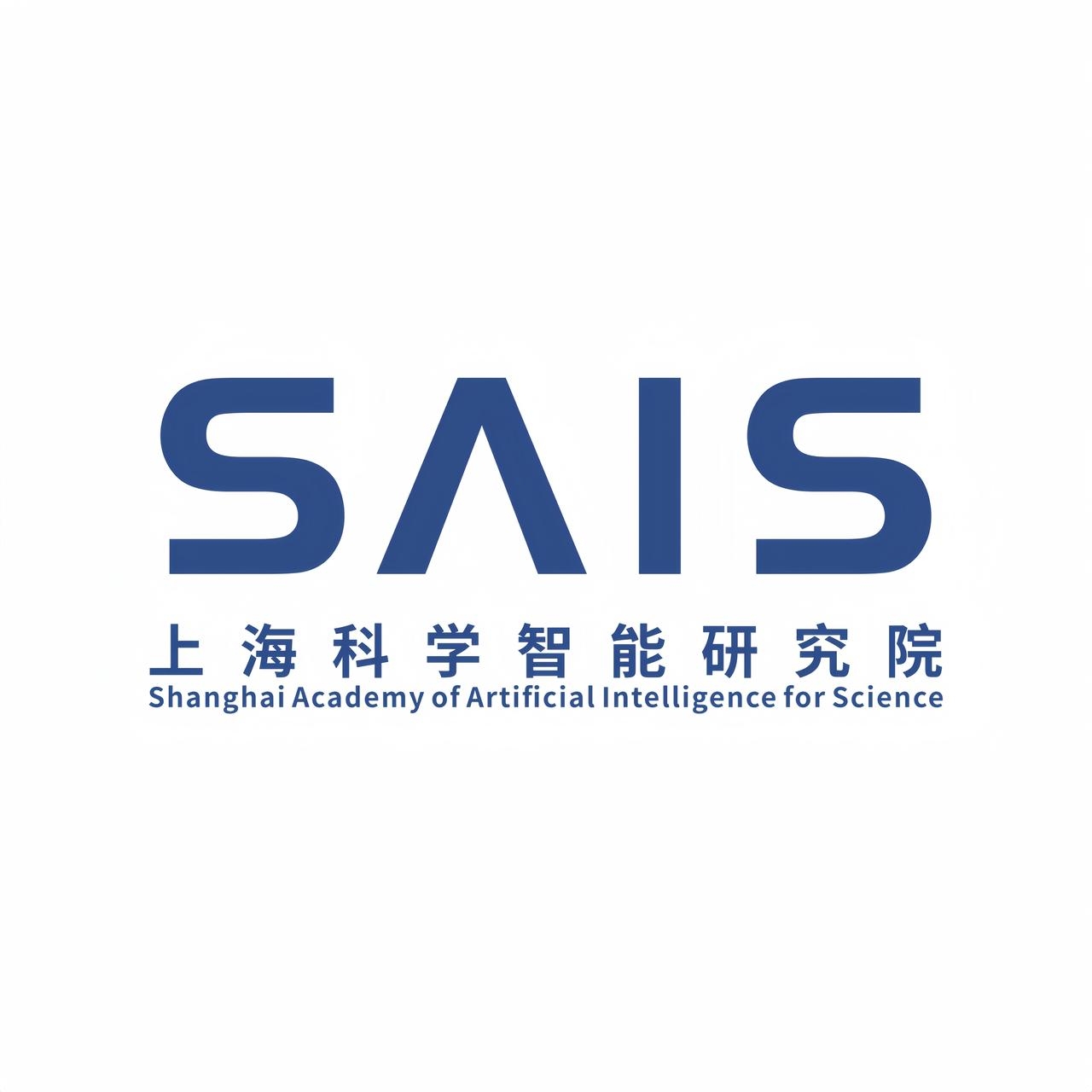}

    \vspace{-2em}
    \hrule
    \vspace{2em}

    \centering
    \raisebox{-0.4\height}{%
      \includegraphics[width=0.1\textwidth]{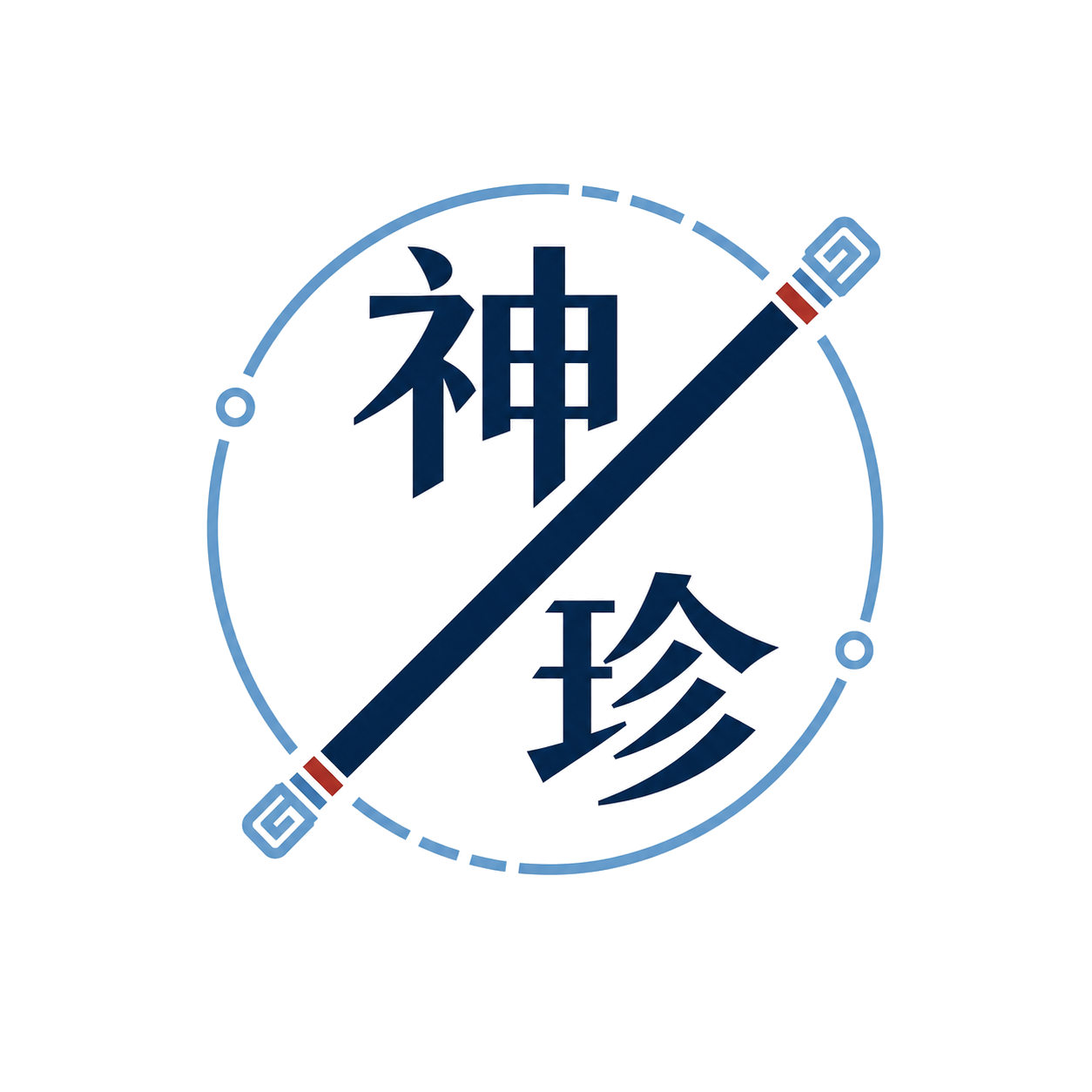}%
    }%
    \hspace{0.2em}%
    \textbf{Monkey King Bang: A Unified Scientific Multimodal Foundation Model}
  \end{minipage}
}

\author{\textbf{SAIS Team, Shanghai Academy of AI for Science}}
\date{}
 
\begin{document}
% \noindent
% \includegraphics[width=0.16\textwidth]{fig/sais.jpeg}

\maketitle

% \suzy{Todo: 1. Adding author list }\\

\begin{abstract}
Scientific discovery is increasingly shifting from isolated disciplines to multi-domain reasoning, and AI for science faces a similar transition. Existing systems are either specialised for individual domains or unify scientific data mainly through text tokenisation and prompt-based interfaces, limiting their ability to handle diverse scientific inputs, produce modality-native outputs, and support joint understanding, reasoning, and generation across scientific domains. 
We introduce \model, a unified scientific multimodal model for both understanding and generation, built around a shared Transformer backbone and modality-tailored encoders, adapters, and decoders. \model\ covers six scientific branches, including DNA, RNA, proteins, small molecules, earth science, and medical images, and supports native outputs such as biological sequences, molecular strings, meteorological fields, and segmentation masks. Training follows a two-stage modality-then-language curriculum: Stage~1 aligns modality-specific components with the frozen backbone, and Stage~2 consolidates them with the language backbone using mixed scientific and general corpora. Experiments show that \model\ achieves competitive scientific understanding across biological and molecular benchmarks, produces high-fidelity native outputs for weather forecasting, biological generation, and medical-image segmentation, and largely retains the general capabilities of its Qwen3-VL backbone. These results demonstrate the feasibility of the proposed paradigm, suggesting that shared-backbone models with modality-tailored components can provide a promising foundation for future cross-domain scientific multimodal exploration. The model and code are publicly available at \url{https://github.com/Shanghai-Academy-of-AI-For-Science/MKB} and \url{https://huggingface.co/sais-org/MKB}.

\end{abstract}

\section{Introduction}
\label{sec:intro}
Scientific discovery increasingly extends beyond the boundaries of individual disciplines. Understanding how genomic variation contributes to disease, for example, often requires integrating evidence from genetics, transcriptomics, protein biology, and molecular pharmacology. Similar interdependencies are pervasive across science, where relevant knowledge is distributed across heterogeneous representations, physical scales, and disciplinary domains. Modern scientific discovery therefore increasingly depends on the coordinated analysis of multiple forms of scientific evidence.

Artificial intelligence for science has followed a similar trajectory. Early AI systems were largely developed as specialist models, each tailored to the data structure and prediction targets of an individual field, and such systems now define the state of the art within many individual scientific domains. These specialist systems can be broadly grouped into two families. The first targets scientific data whose structure resembles common visual or spatio-temporal signals. For medical images and atmospheric fields, segmentation and forecasting models adapt vision or spatio-temporal architectures and learn domain-specific structure from large modality-native corpora, producing sharper segmentation masks and lower forecast errors than domain-agnostic backbones~\cite{sam3,biomedparse,chen2023fuxi,pangu-weather,fourcastnet,graphcast}.
The second targets scientific objects whose structure is difficult to capture through generic visual or textual representations. Proteins, DNA, and RNA require specialised alphabets and long-range sequence modelling, while molecules are more naturally represented as graphs or chemically meaningful tokenisations. These requirements have led to strong specialist models, including protein language models~\cite{esm1v,esm2}, nucleotide encoders~\cite{dnabert2_gue,nucleotide_transformer}, and molecular graph networks~\cite{suiren}. Their empirical strength arises from architectural priors and modality-native supervision that a generic backbone would otherwise need to learn from scratch. By construction, however, these systems typically remain confined to individual domains and do not support direct cross-domain composition within a single model.

Recent scientific generalists partly address cross-domain modelling by connecting diverse scientific inputs to a shared language backbone, typically through text-like serialisation or tool-mediated prompts~\cite{naturelm,biology_instructions,intern_s1_pro}. 
Such approaches provide a unified conversational interface, but their predominantly text-centric interaction mechanisms are not equally suitable for all scientific data. For inputs, dense atmospheric fields, molecular structures, and biomedical images contain spatial, geometric, or numerical structure that may be difficult to preserve through serialisation. 
For outputs, tasks such as weather forecasting and medical-image segmentation require native dense predictions, including latitude--longitude fields and per-pixel masks, while text-centric models can only describe these outputs or delegate them to external predictors, rather than decoding them directly from shared hidden representations.
Even the largest current scientific multimodal LLMs, such as the trillion-parameter Intern-S1-Pro~\cite{intern_s1_pro}, focus on textual scientific reasoning and do not natively emit continuous-field forecasts or high-resolution segmentation masks.

\textbf{Taken together, existing systems leave an important gap: a single shared-backbone model that can directly encode heterogeneous scientific modalities in structurally appropriate forms, jointly contextualise information across them, and generate modality-native outputs across multiple scientific domains.}

Motivated by this gap, we introduce \underline{M}onkey \underline{K}ing \underline{B}ang (\model), a unified scientific multimodal foundation model for understanding and generating heterogeneous scientific data.
\model\ covers six scientific branches: DNA, RNA, proteins, small molecules, Earth-system data, and medical images. It is built around a shared autoregressive Transformer backbone, with modality-tailored encoders, adapters, and, where applicable, decoders.
For a given task, the relevant encoder--adapter pairs map scientific inputs into modality-token sequences aligned with the hidden space of the shared backbone. These modality blocks are arranged according to the task and jointly contextualised within a shared sequence. For generation tasks, the target-modality hidden states are selected from the contextualised sequence and passed to the corresponding domain-specific decoder, which produces outputs in the required native form. This design allows the modalities involved in a task to interact through a shared representation space while retaining important modality-specific properties, including sequential dependencies, molecular geometry, continuous physical signals, and high-dimensional spatial structure.

However, jointly training heterogeneous modality-specific pathways within a shared backbone is challenging, as differences in data structure, scale, and supervision can lead to unstable optimisation and interference between modalities. To address this, training follows a two-stage modality-then-language curriculum. 
In the first stage, each modality-specific pathway is aligned with the frozen shared backbone, enabling its encoder, adapter, and, when applicable, decoder to establish a stable interface with the backbone representation space.
In the second stage, the modality components and language backbone are jointly consolidated using mixed scientific and general-purpose corpora. This curriculum integrates heterogeneous scientific capabilities into a single checkpoint while limiting degradation of the backbone's original language and vision capabilities.

\begin{figure}[t]
    \centering
    \includegraphics[width=0.95\textwidth]{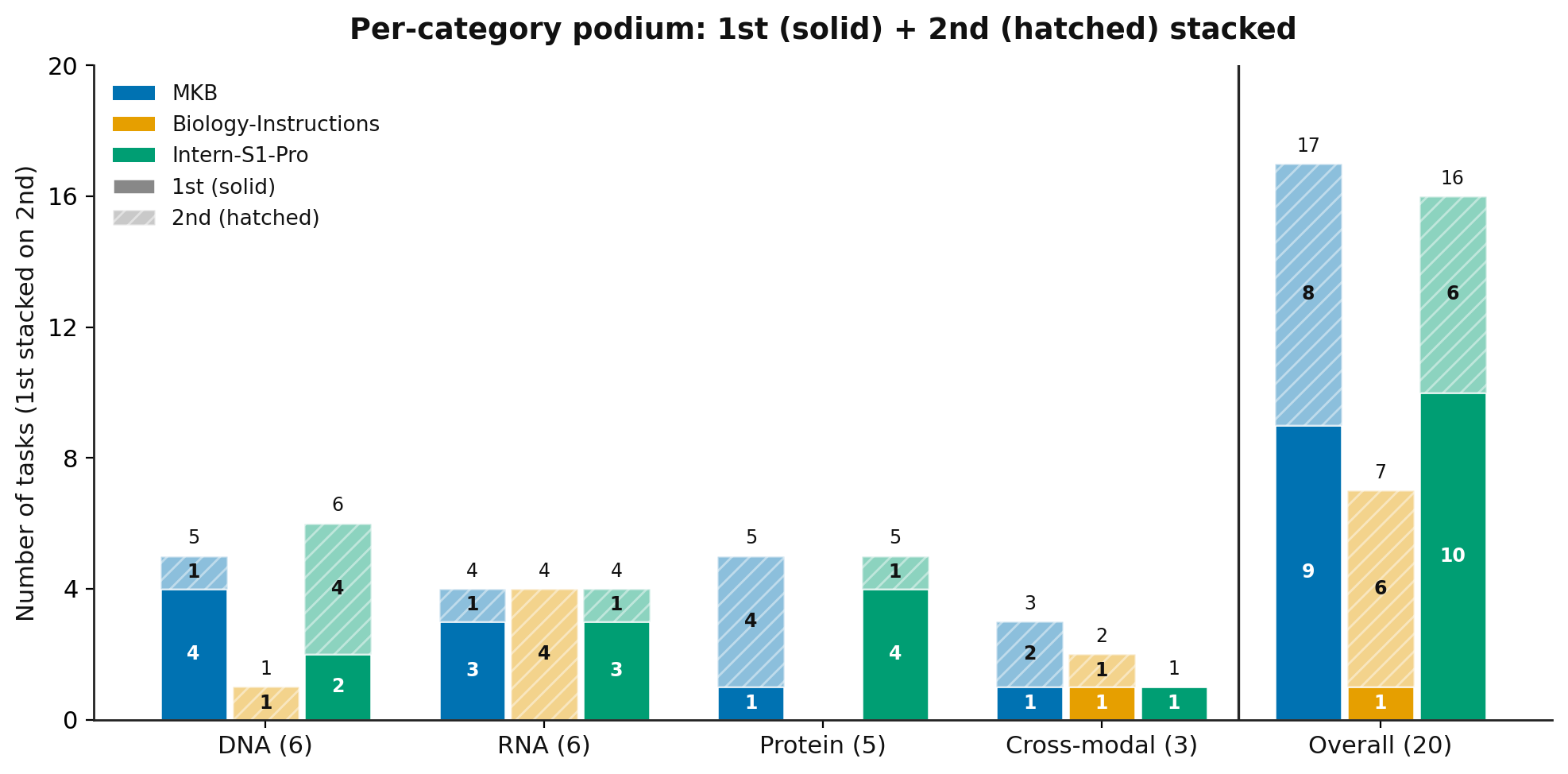}
    \caption{\textbf{Podium coverage on the Biology-Instructions understanding suite.} For each biological category and the 20-task overall, bars stack first-place (solid) on second-place (hatched) finishes, and the top-of-bar number is their sum. \model\ (11B) reaches $17/20$ top-two, ahead of Intern-S1-Pro (1T, $16/20$) and far above Biology-Instructions (8B, $7/20$).}
    \label{fig:podium}
\end{figure}

We evaluate \model\ along three complementary dimensions: scientific understanding, scientific generation, and general capabilities. Across biological sequence understanding benchmarks, \model\ ranks within the top two on $17/20$ tasks, surpassing the LLM-based state-of-the-art systems ($16/20$)~\cite{intern_s1_pro} despite using roughly two orders of magnitude fewer parameters (Figure~\ref{fig:podium}). On molecular benchmarks, it also surpasses specialist models on a subset of tasks. For native scientific generation, it produces high-fidelity outputs across biological sequences, molecular strings, meteorological fields, and medical-image segmentation masks. In particular, \model\ achieves strong Earth-system forecasting performance, including better medium-range forecasts than HRES~\cite{haiden2021ecmwf} on the evaluated settings, and obtains the best overall Dice performance across the evaluated medical-imaging modalities. After joint scientific multimodal training, it also largely retains the general-purpose capabilities of its Qwen3-VL backbone~\cite{qwen3vl}.

Overall, these results demonstrate the feasibility of a shared-backbone modelling paradigm in which biological sequences, molecular graphs, spatial image features, and continuous physical fields are encoded into a common representation space and decoded into modality-native scientific outputs. They also highlight remaining challenges in precise scalar regression, particularly for ADMET prediction, where stronger quantitative supervision and regression-oriented modelling may be beneficial.

\newcommand{\stoken}[1]{\texttt{<|#1\_start|>}}
\newcommand{\etoken}[1]{\texttt{<|#1\_end|>}}
\newcommand{\ptoken}[1]{\texttt{<|#1\_pad|>}}
\newcommand{\R}{\mathbb{R}}
\newcommand{\BQ}{\textsc{Bio-Qwen3-VL}}

\section{Architecture}
\label{sec:arch}

\begin{figure}[!t]
  \vskip 0.2in
  \begin{center}
    \centerline{\includegraphics[width=\columnwidth]{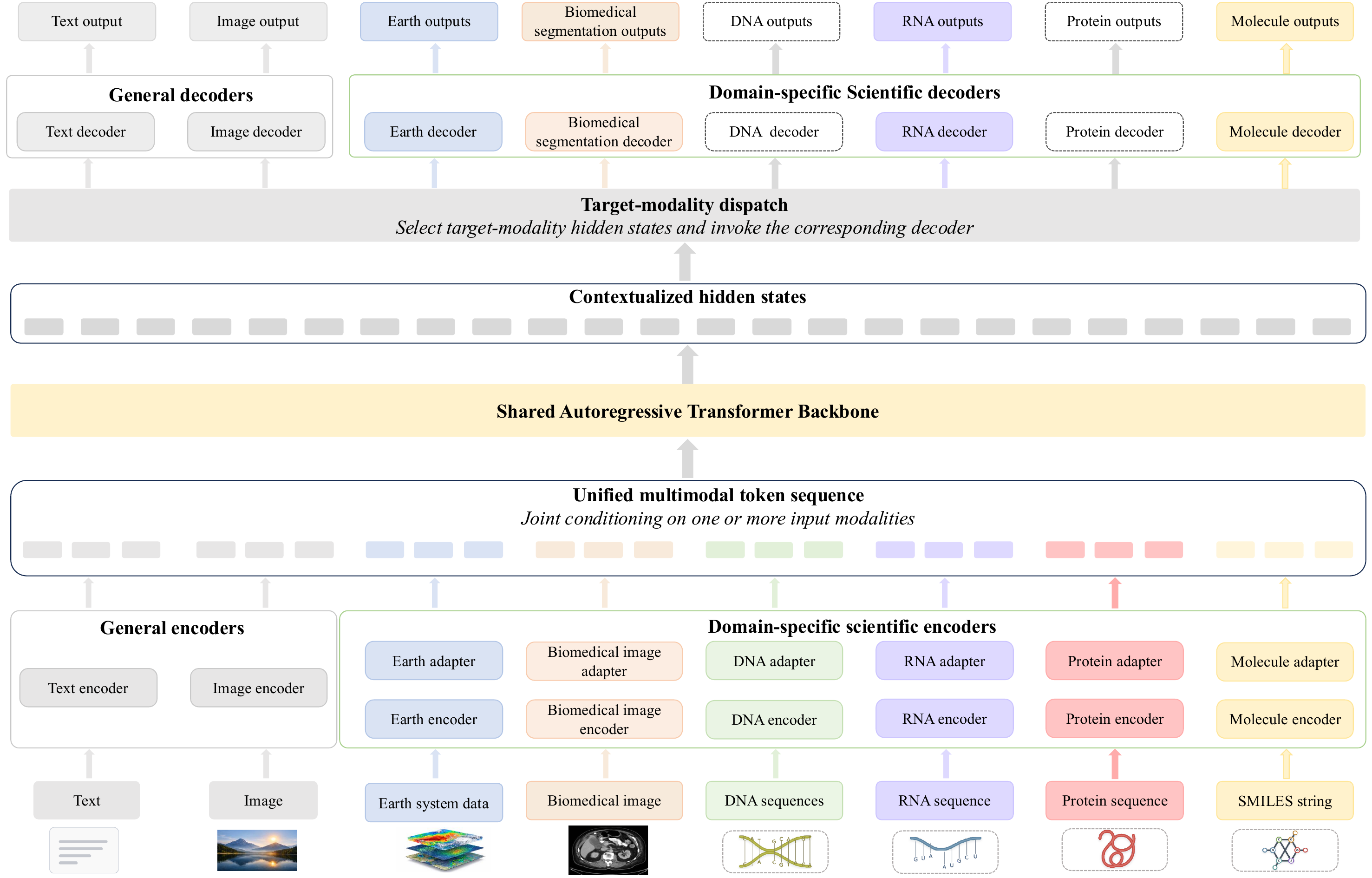}}
    \caption{Overview of \model. \model\ consists of general and domain-specific encoders, adapters, and decoders attached to a shared autoregressive Transformer backbone. Heterogeneous inputs are mapped into a unified multimodal token sequence for joint contextualization, after which target-modality hidden states are dispatched to the corresponding decoder to generate outputs in their native forms. Dashed pathways denote framework components that are not evaluated for native generation in this report.}
    \label{fig:framework}
  \end{center}
  \vspace{-0.8cm}
\end{figure}

\subsection{Overview}
We propose \model, a unified scientific multimodal model that supports scientific understanding and modality-native generation across heterogeneous scientific domains. The model is built around a shared \textbf{Qwen3-VL-8B} Transformer backbone~\cite{qwen3vl}, with hidden dimension ($D_{\mathrm{LLM}}$=4096), together with structure-aware representation pathways and modality-specific generation components. \model\ covers six scientific branches: DNA, RNA, proteins, small molecules, Earth-system data, and medical images, while retaining the general capabilities of Qwen3-VL.

Scientific inputs are represented according to their native structural properties before being aligned with the hidden space of the shared backbone. Biological sequences and molecular graphs are encoded into variable-length features and compressed into fixed-length latent tokens through Perceiver-style resamplers and modality-specific projectors. Earth-system fields instead retain a dense latitude--longitude token grid to preserve their spatial organisation, while biomedical segmentation uses a dual-path design that combines instruction-conditioned Qwen3-VL representations with dense SAM3 image features. For each task, only the relevant modalities are composed into the shared multimodal sequence and jointly contextualised by the autoregressive Transformer.

For understanding tasks, the shared language-model head generates textual responses autoregressively. For generation tasks, hidden states associated with the target modality are selected from the contextualised sequence and passed to the corresponding generation pathway, which produces outputs in their native form, including biological sequences, molecular strings, meteorological fields, and segmentation masks. The following sections describe the unified representation framework, the modality-specific representation pathways, the shared backbone and multimodal composition mechanism, and the modality-native generation pathways.

\begin{table}[t]
\centering
\caption{
Summary of the modality-specific representation and generation pathways.
All branches support textual understanding through the shared language-model head.
For sequence and graph modalities, encoder features are converted by a modality adapter, consisting of a Perceiver-style resampler and an MLP projector, into the shared Transformer hidden size $D_{\mathrm{LLM}}=4096$.
Dashes indicate that no modality-native generation pathway is used in the reported experiments.
}
\label{tab:modules}
\small
\renewcommand{\arraystretch}{1.12}
\resizebox{\textwidth}{!}{%
\begin{tabular}{
L{1.7cm}
L{2.1cm}
L{3.1cm}
L{4.0cm}
L{3.3cm}
L{3.5cm}
}
\toprule
\textbf{Branch} &
\textbf{Native input} &
\textbf{Representation pathway} &
\textbf{Key specification} &
\textbf{Backbone-facing representation} &
\textbf{Native-output pathway} \\
\midrule

DNA &
Nucleotide sequence &
Encoder + modality adapter &
1-D convolutional Transformer; width $512$; $8$ blocks; $K=64$ &
$64$ tokens in $\mathbb{R}^{D_{\mathrm{LLM}}}$ &
-- \\

\addlinespace

RNA &
Nucleotide sequence &
Encoder + modality adapter &
1-D convolutional Transformer; width $512$; $8$ blocks; $K=64$ &
$64$ tokens in $\mathbb{R}^{D_{\mathrm{LLM}}}$ &
Linear nucleotide head; RNA sequence \\

\addlinespace

Protein &
Amino-acid sequence &
Encoder + modality adapter &
ESM2-150M; width $640$; $K=64$ &
$64$ tokens in $\mathbb{R}^{D_{\mathrm{LLM}}}$ &
-- \\

\addlinespace

Small molecule &
Molecular graph &
Encoder + modality adapter &
Suiren-ConfAvg; width $256$; local and fully connected graphs; $K=64$ &
$64$ tokens in $\mathbb{R}^{D_{\mathrm{LLM}}}$ &
Cross-attentive Transformer; 226-token SMILES vocabulary \\

\addlinespace

Earth system &
Dense physical field &
Structure-preserving spatial projection &
Swin-style encoder; width $2048$; $12$ blocks; $6\times6$ patches &
Dense $120\times240$ latitude--longitude token grid &
Swin-style field decoder; 70-channel meteorological field \\

\addlinespace

Biomedical image &
Image and text instruction &
Dual-path semantic and spatial representation &
Qwen3-VL semantic pathway; SAM3 dense-image pathway; $2016\times2016$ input &
Semantic conditioning tokens; dense SAM3 features retained for decoder &
SAM3 mask decoder; $576\times576$ segmentation mask \\

\bottomrule
\end{tabular}
}
\end{table}

\subsection{Unified Scientific Representation Framework}
For each task, the participating scientific modalities are processed through modality-specific representation pathways that preserve their native structural properties while aligning them with the hidden space of the shared Transformer. Given a native input $\mathbf{X}_{(m)}$ from modality $m$, the corresponding encoder produces continuous representations
\begin{equation}
\mathbf{H}_{(m)}
\in
\mathbb{R}^{N_{(m)} \times d_{(m)}},
\end{equation}
where $N_{(m)}$ and $d_{(m)}$ denote the number and dimensionality of the encoder features, respectively.

For sequence-like and graph-like modalities, including DNA, RNA, proteins, and molecules, the encoder output is passed to a modality-specific adapter. Each adapter consists of a Perceiver-style resampler followed by a lightweight MLP projector, thereby converting variable-length encoder features into a fixed number of backbone-facing tokens. Specifically, the resampler uses learnable latent queries
$\mathbf{Q}_{(m)}\in\mathbb{R}^{K_{(m)} \times d_{(m)}}
$
to attend to the encoder features and produce
\begin{equation}
\mathbf{Z}_{(m)}
=
\operatorname{Resampler}_{(m)}
\left(
\mathbf{H}_{(m)};
\mathbf{Q}_{(m)}
\right)
\in
\mathbb{R}^{K_{(m)} \times d_{(m)}}.
\end{equation}
Here, $K_{(m)}$ denotes the number of latent tokens, with all sequence and graph pathways using $K_{(m)}=64$ in the current implementation. The MLP projector then aligns the resampled representations with the hidden space of the shared Transformer:
\begin{equation}
\mathbf{F}_{(m)}
=
\operatorname{MLP}_{(m)}
\left(
\mathbf{Z}_{(m)}
\right)
\in
\mathbb{R}^{K_{(m)} \times D_{\mathrm{LLM}}}.
\end{equation}
Together, the resampler and MLP form a compact modality adapter that controls the number of tokens introduced into the shared backbone while preserving the sequence- or graph-level information extracted by the native encoder.

Dense scientific modalities instead use structure-preserving representation pathways. For Earth-system data, the spatial encoder retains the complete latitude--longitude patch grid and projects the resulting weather tokens directly into the shared hidden space, avoiding the fixed-length latent bottleneck used for sequence and graph modalities. This preserves the two-dimensional spatial organisation required by the positional encoding and field decoder. Biomedical segmentation adopts a dual-path design in which the Qwen3-VL image--text pathway provides instruction-conditioned semantic representations, while a parallel SAM3~\cite{sam3} vision pathway preserves high-resolution spatial features for mask prediction.

For each task, only the relevant input and target modalities are instantiated. Their backbone-facing representations are arranged within a shared multimodal token sequence and jointly contextualised by the autoregressive Transformer. For understanding tasks, the contextualised representations support autoregressive textual prediction. For generation tasks, the hidden states associated with the target modality are selected and passed to the corresponding modality-specific decoder. Where applicable, the Earth-system and biomedical pathways additionally retain native dense features that support the reconstruction of spatially detailed outputs.

\subsection{Modality-specific Input Representation Pathways}
\label{subsec:input_pathway}
\subsubsection{Biological Sequence Representation}
Biological sequences are represented using sequence encoders that preserve local motifs and long-range dependencies before mapping variable-length inputs into the fixed-length latent representation described above. We use separate pathways for nucleotide and protein sequences because their alphabets, structural patterns, and available pretrained representations differ.

\paragraph{Nucleotide sequences.}
For DNA and RNA, the inputs are character-level nucleotide sequences,
$
\mathbf{X}_{\mathrm{DNA}}
\in
\left\{\mathrm{A},\mathrm{T},\mathrm{G},\mathrm{C},\mathrm{N}\right\}^{L},
$
and
$
\mathbf{X}_{\mathrm{RNA}}
\in
\left\{\mathrm{A},\mathrm{U},\mathrm{G},\mathrm{C},\mathrm{N}\right\}^{L},
$, respectively.
Here, $L$ denotes the variable sequence length and $\mathrm{N}$ represents an ambiguous nucleotide. DNA and RNA are tokenised at the character level using separate modality-specific vocabularies, $\mathcal{V}{\mathrm{DNA}}$ and $\mathcal{V}{\mathrm{RNA}}$. For RNA inputs containing thymine, $\mathrm{T}$ is normalised to uracil, $\mathrm{U}$, before tokenisation. Both DNA and RNA sequences are truncated to at most 2048 tokens.

DNA and RNA are encoded by independently parametrised one-dimensional convolutional Transformers trained on their respective nucleotide corpora. Each encoder combines a convolutional stem for capturing local motif-level patterns with Transformer blocks for modelling longer-range dependencies, producing contextualised per-token representations
$
\mathbf{H}_{\mathrm{DNA}}
\in
\mathbb{R}^{B \times L'_{\mathrm{DNA}} \times d_{\mathrm{DNA}}},
$
and
$
\mathbf{H}_{\mathrm{RNA}}
\in
\mathbb{R}^{B \times L'_{\mathrm{RNA}} \times d_{\mathrm{RNA}}}
$, respectively.
The two encoders share the same architectural design but use modality-specific vocabularies and independent parameters. Their detailed configurations are summarised in Table~\ref{tab:modules}.

\paragraph{Protein sequences.}
For proteins, we use the pretrained ESM2 encoder~\cite{esm2} to extract residue-level representations from an amino-acid sequence
$\mathbf{X}_{\mathrm{prot}} \in \Sigma^{L}$,
where $\Sigma$ denotes the ESM2 amino-acid vocabulary. Sequences are truncated to at most $1024$ residues and processed by ESM2 to produce contextualised residue embeddings
$
\mathbf{H}_{\mathrm{prot}}
\in
\mathbb{R}^{B \times L'_{\mathrm{prot}} \times d_{\mathrm{prot}}}.
$
Unlike the nucleotide encoders, which are trained specifically for \model, the protein pathway builds on the pretrained sequence representations learned by ESM2.

The resulting DNA, RNA, and protein features are passed through their respective modality adapters, each consisting of a Perceiver-style resampler followed by an MLP projector, producing $64$ backbone-facing tokens for each participating biological modality.

\subsubsection{Molecular Graph Representation}
Small molecules are represented as molecular graphs in order to preserve their native chemical connectivity. Given a SMILES string $\mathbf{X}_{\mathrm{mol}}$, we parse it into a two-dimensional graph
$
\mathcal{G}=(\mathcal{V},\mathcal{E}),
$
where $\mathcal{V}$ and $\mathcal{E}$ denote the sets of atoms and chemical bonds, respectively. Atom and bond attributes encode standard chemical properties, including atomic identity, charge, hybridisation, aromaticity, bond type, and stereochemistry.

To capture both local chemical structure and longer-range atom--atom dependencies, the molecular representation uses two complementary connectivity patterns: a sparse local graph $\mathbf{A}_{\mathrm{loc}}$, containing the observed chemical bonds, and a fully connected graph $\mathbf{A}_{\mathrm{full}}$ over all atom pairs. The local graph preserves bonded interactions and functional-group structure, while the fully connected graph enables information exchange between chemically distant atoms.

Atom-level representations are extracted using the pretrained Suiren-ConfAvg graph foundation model~\cite{suiren}. Given $\mathcal{G}$, Suiren produces contextualised atom embeddings
$
\mathbf{H}_{\mathrm{mol}}
\in
\mathbb{R}^{B \times N_{\mathrm{atom}} \times d_{\mathrm{mol}}},
$
where $N_{\mathrm{atom}}$ denotes the number of atoms after batch padding. The resulting atom features are passed through the molecule-specific modality adapter, consisting of a Perceiver-style resampler followed by an MLP projector, producing $64$ backbone-facing molecular tokens.

\subsubsection{Dense Physical-field Representation}
Earth-system data are represented as dense physical fields in order to preserve their continuous spatial organisation. The input is a global gridded atmospheric state
$\mathbf{X}_{\mathrm{w}}\in\mathbb{R}^{B\times T_{\mathrm{in}}\times C\times H\times W}$, where $T_{\mathrm{in}}=1$, $C=70$, and $(H,W)=(721,1440)$ correspond to the latitude--longitude dimensions of the $0.25^{\circ}$ grid. The input channels include upper-air and surface meteorological variables, while static geographical fields and temporal metadata provide additional spatial and forecast conditioning.

A Swin-style shifted-window Transformer encodes the input into a dense latitude--longitude patch grid. Using a patch size of ($6\times6$), the encoder produces
$(H_{\mathrm{p}},W_{\mathrm{p}})=(120,240)$, yielding $28{,}800$ weather tokens.
The encoder configuration is summarised in Table~\ref{tab:modules}.
Unlike sequence-like modalities, these dense meteorological tokens are not compressed by a resampler. Instead, they are projected to the shared Transformer hidden space through the cross-modal merger and inserted into reserved atmosphere placeholder positions. This preserves the two-dimensional latitude--longitude structure required by the spatial positional encoding and the downstream field decoder.

\subsubsection{Dual-path Biomedical Image Representation}
Biomedical image segmentation requires both semantic understanding of the instruction and preservation of fine-grained spatial details. We therefore adopt a dual-path representation design that processes the same medical image through complementary semantic and spatial pathways.

Given a medical image $I$ and a text instruction $p$, the semantic pathway reuses the native Qwen3-VL image--text encoding pipeline. The image and instruction are encoded as a multimodal sequence, and the shared Transformer produces instruction-conditioned hidden states $\mathbf{H}_{\mathrm{sem}}\in\mathbb{R}^{B\times L_{\mathrm{sem}}\times D_{\mathrm{LLM}}}$. These hidden states are projected into semantic conditioning tokens for the segmentation decoder.

In parallel, the spatial pathway processes the same image with the SAM3 image processor and vision backbone~\cite{sam3}, producing dense visual features that preserve local boundaries and fine-grained spatial structure. In our experiments, input images are resized to $2016\times2016$, and the segmentation pathway predicts masks at a spatial resolution of $576\times576$.

Together, these complementary pathways provide the segmentation decoder with instruction-aware semantic context and dense spatial evidence, enabling accurate pixel-level biomedical mask prediction.

\subsection{Shared Backbone and Multimodal Composition}
The shared backbone of \model\ is initialised from Qwen3-VL-8B~\cite{qwen3vl}, which provides a pretrained vision--language representation space and a decoder-only causal Transformer for multimodal contextualisation. The language Transformer contains $36$ layers with hidden size $D_{\mathrm{LLM}}=4096$. We retain its native attention, normalisation, feed-forward, and multi-axis rotary positional encoding mechanisms, as well as the original vision tower for processing general images and videos.

For each task, text tokens, optional Qwen3-VL image or video tokens, and the backbone-facing representations of the participating scientific modalities are arranged into a single multimodal sequence,
$
\mathbf{E}_{\mathrm{fused}}
\in
\mathbb{R}^{B \times L_{\mathrm{tot}} \times D_{\mathrm{LLM}}},
$
where $L_{\mathrm{tot}}$ depends on the modalities involved in the task. Scientific representations are inserted at reserved modality-specific positions delimited by dedicated control tokens. Sequence and graph modalities contribute fixed-length adapter tokens, whereas dense Earth-system representations retain their latitude--longitude token layout. The resulting sequence is processed jointly by the causal Transformer, enabling information exchange among the modalities participating in the current task.

For textual understanding tasks, the contextualised hidden states are decoded autoregressively through the native language-model head. For modality-native generation, control tokens and reserved target positions identify the required output pathway. The hidden states associated with the target modality are selected from the contextualised sequence and passed to the corresponding generation component, such as the RNA head, molecular decoder, physical-field decoder, or biomedical segmentation decoder. Molecular generation uses a specialised SMILES vocabulary that remains separate from the language-model vocabulary.

Therefore, the shared backbone provides a common contextual representation space across heterogeneous scientific inputs, while modality-specific generation pathways preserve the structural requirements of their respective outputs. For dense Earth-system and biomedical outputs, the shared hidden states additionally condition native spatial features retained by their structure-preserving representation pathways.

\subsection{Modality-native Output Generation Pathways}
\label{subsec:output-generation}
As summarised in Table~\ref{tab:modules}, \model\ supports modality-native generation for RNA sequences, small molecules, Earth-system fields, and biomedical segmentation masks. Given contextualised backbone states $\mathbf{H}_{\mathrm{LLM}}\in\mathbb{R}^{B\times L_{\mathrm{tot}}\times D_{\mathrm{LLM}}}$, the hidden states associated with the target modality are selected from the shared multimodal sequence and passed to the corresponding output pathway. These output components differ according to the structural requirements of each target modality.

\paragraph{RNA sequence generation.}
Because the shared backbone already provides autoregressive contextualisation across the target sequence, RNA generation uses a lightweight bias-free linear head. The head projects the target-modality hidden states $\mathbf{H}^{\mathrm{RNA}}_{\mathrm{target}}\in\mathbb{R}^{B\times L_{\mathrm{RNA}}\times D_{\mathrm{LLM}}}$ directly into the RNA nucleotide vocabulary. RNA tokens are generated autoregressively, and the head is trained using standard shifted cross-entropy.

\paragraph{Molecular generation.}
Molecular generation uses an independent autoregressive Transformer because the specialised whole-atom SMILES vocabulary is decoupled from the language-model vocabulary. Hidden states gathered from the molecular conditioning positions, $\mathbf{H}^{\mathrm{mol}}_{\mathrm{target}}\in\mathbb{R}^{B\times L_{\mathrm{cond}}\times D_{\mathrm{LLM}}}$, are provided as cross-attention context to a Flamingo-style Transformer decoder~\cite{alayrac2022flamingo}. The decoder attends causally to the generated SMILES prefix while cross-attending to the shared multimodal context, and predicts logits $\widehat{\mathbf{Y}}_{\mathrm{mol}}\in\mathbb{R}^{B\times L_{\mathrm{mol}}\times V_{\mathrm{mol}}}$ over a specialised $V_{\mathrm{mol}}=226$ whole-atom SMILES vocabulary. It is trained with shifted cross-entropy. At inference time, the generated whole-atom tokens are detokenised into a SMILES string.

\paragraph{Dense physical-field generation.}
The Earth-system output pathway converts the contextualised weather-token states back into a dense meteorological field. A back-projection module maps the selected weather states $\mathbf{H}^{\mathrm{w}}_{\mathrm{target}}\in\mathbb{R}^{B\times L_{\mathrm{w}}\times D_{\mathrm{LLM}}}$ to the native weather-token dimension $d_{\mathrm{w}}$. The projected features are fused with encoder skip features and refined by a Swin-style shifted-window Transformer decoder. The resulting token grid is reshaped into a two-dimensional feature map and upsampled to the original $0.25^{\circ}$ resolution, yielding a one-step prediction $\widehat{\mathbf{Y}}_{\mathrm{earth}}$. Multi-step forecasts are generated by autoregressive rollout, where each predicted field is re-fed as the next encoder input together with the updated lead-time embedding.

The Earth-system decoder is trained with a latitude-weighted Charbonnier loss, using cosine latitude weights to account for grid-cell area and channel weights to balance variables with different dynamic ranges. Specific humidity is down-weighted with $\alpha_{\mathrm{hum}}=0.3$, near-surface temperature and 10 m winds use $\alpha_{\mathrm{sur}}=0.5$, and upper-air variables together with mean sea-level pressure use $\alpha_{\mathrm{upper}}=1.0$. During training, rollout supervision is applied only at selected checkpoints, including the verification step, which reduces memory consumption while still exposing the model to autoregressive error accumulation. During inference, the same autoregressive rollout procedure is used to generate forecasts at user-specified horizons.

\paragraph{Biomedical mask generation.}
Biomedical segmentation combines the two complementary representations introduced in the dual-path image pathway. The contextualised Qwen3-VL states provide instruction-conditioned semantic information, while the SAM3 image branch supplies dense multi-scale spatial features. The SAM3 mask decoder~\cite{sam3} fuses these representations and predicts object-level classification scores, bounding boxes, and segmentation masks $\{(\ell_q,\mathbf{b}_q,\mathbf{M}_q)\}_{q=1}^{Q}$, where $Q$ denotes the number of object queries.

The segmentation pathway is trained with standard SAM3-style multi-objective supervision. Object-query predictions are matched to ground-truth instances through Hungarian assignment using classification, box, and mask-overlap costs. The matched predictions are optimized with sigmoid focal classification loss, L1 box loss, generalized IoU loss, sigmoid focal mask loss, and mask Dice loss. The reported configuration further introduces an image-level semantic-mask Dice objective and an auxiliary meta-object cross-entropy objective. At inference time, the highest-confidence mask is upsampled to the target resolution as the final instruction-guided segmentation output.

\section{Data Construction}
\label{sec:data}
\model\ is trained on six scientific modalities together with scientific text, general text, and vision--language data. Because scientific supervision is scarce, noisy, and distributed across incompatible file formats, we treat data construction as a first-class component of the system. This section describes the coverage of scientific tasks and modalities, the unified instruction schema for native objects, the construction and leakage control of cross-modal supervision, and the Stage-2 consolidation mixture.

\subsection{Scientific Data Coverage}
\label{sec:data-sources}
For \textbf{biological sequences and molecules}, we primarily adopt established benchmarks to maintain comparability with prior work.
DNA, RNA, protein, and RNA--protein tasks are primarily drawn from Biology-Instructions~\cite{biology_instructions}, covering sequence understanding, regulatory prediction, protein-property prediction, and biological interaction tasks. Molecular data cover property prediction, ADMET assessment, molecule understanding, and text-conditioned molecular generation, primarily using SMolInstruct~\cite{smolinstruct}, TDC~\cite{tdc}, and MoleHB~\cite{suiren}.
\textbf{Earth-system} forecasting uses ERA5 reanalysis~\cite{era5} at $0.25^{\circ}$ resolution, and \textbf{medical-image segmentation} uses BiomedParse~\cite{biomedparse} across nine imaging modalities. General multimodal capability is maintained with the native image--text, OCR, grounding, and instruction data inherited from the Qwen3-VL~\cite{qwen3vl} backbone. Native objects are standardized, and invalid records are removed before training.

\subsection{Unified instruction schema}
\label{sec:data-schema}
A central design choice is that native modality objects are \emph{not tokenized as ordinary prompt text}. Instead, native objects are stored separately from the dialogue and linked to it through modality references.
The dialogue refers to each object through a reserved modality placeholder, such as \texttt{<protein>}.
At data-loading time, each placeholder is expanded into the corresponding encoder segment (Section~\ref{sec:arch}), allowing the language model to condition on continuous modality representations rather than lossy textual serializations. This schema keeps the interface consistent across understanding, generation, and cross-modal samples.

Each sample is formatted as an instruction dialogue: the system turn specifies the task and output contract, the human turn contains the placeholders and question, and the assistant turn contains the supervised answer. Only assistant answer tokens are supervised by the text loss, while prompt tokens, modality placeholders, and decoder-output positions are masked. For native-generation tasks, an output placeholder in the assistant response routes the corresponding hidden states to the target-modality decoder rather than the language head. Continuous targets are normalized using training-set statistics only.

\subsection{Cross-modal Data Construction}
\label{sec:data-crossmodal}
Ready-made instruction data is dominated by single-modality tasks, whereas many scientific questions require reasoning across coupled biological and chemical entities. We therefore build cross-modal datasets directly from authoritative, openly redistributable databases, as summarized in Table~\ref{tab:crossmodal-data}. 
A source-agnostic intermediate representation standardizes entities, relations, provenance, and evidence across databases.
Our pipeline is built around a Common Intermediate Representation (CIR) that standardizes heterogeneous database records into typed scientific entities, relations, and associated provenance and evidence. By decoupling source-specific parsing from downstream task construction, the CIR enables consistent entity reuse across datasets, simplifies the integration of new data sources, and supports systematic construction of cross-modal tasks.

\paragraph{Enzyme catalysis (protein $\times$ molecule).} 
We construct enzyme-conditioned molecular generation data by linking curated enzyme sequences to reaction substrates, products, and cofactors~\cite{rhea,chebi}.
This construction yields two generation tasks: (1)~substrate-to-product prediction, conditioned on an enzyme sequence and substrate set, and (2)~cofactor prediction, conditioned on an enzyme sequence. 
We retain directionally resolved, mass-balanced reactions and reserve high-confidence annotations for evaluation.

\paragraph{Molecular interaction and binding (protein/RNA $\times$ molecule).} We include drug--target binding-affinity regression from DAVIS and KIBA~\cite{deepdta} and BindingDB~\cite{bindingdb}, with measurements split by affinity type and $z$-scored independently. Together with RNA--protein interaction data from Biology-Instructions~\cite{biology_instructions}, these datasets provide explicit supervision for cross-modal reasoning over biological macromolecules and small molecules.

\begin{table}[t]
  \centering
  \caption{Cross-modal supervision used in training and evaluation. Grouped labels indicate generation (gen), affinity regression (reg), and binary interaction classification (cls).}
  \label{tab:crossmodal-data}
  \begingroup
  \small
  \setlength{\tabcolsep}{4pt}
  \renewcommand{\arraystretch}{1.15}
  \begin{tabularx}{\linewidth}{
    @{}
    >{\centering\arraybackslash}p{0.06\linewidth}
    >{\raggedright\arraybackslash}p{0.28\linewidth}
    >{\raggedright\arraybackslash}p{0.34\linewidth}
    >{\raggedright\arraybackslash}X
    @{}}
    \toprule
    \textbf{Type} & \textbf{Dataset} & \textbf{Modalities} &
    \textbf{Source} \\
    \midrule
    \multirow{2}{*}{gen} &
    \texttt{enzyme\_substrate2product} &
    \texttt{<protein>} \(+\) \texttt{<mol>} \(\rightarrow\) \texttt{<mol>} &
    UniProt+Rhea+ChEBI \\
    & \texttt{enzyme\_cofactor} &
    \texttt{<protein>} \(\rightarrow\) \texttt{<mol>} &
    UniProt+ChEBI \\
    \midrule
    \multirow{6}{*}{reg} &
    \texttt{davis\_dti} &
    \texttt{<protein>} \(+\) \texttt{<mol>} \(\rightarrow\) scalar &
    DeepDTA (DAVIS) \\
    & \texttt{kiba\_dti} &
    \texttt{<protein>} \(+\) \texttt{<mol>} \(\rightarrow\) scalar &
    DeepDTA (KIBA) \\
    & \texttt{bindingdb\_ki} &
    \texttt{<protein>} \(+\) \texttt{<mol>} \(\rightarrow\) scalar &
    BindingDB \\
    & \texttt{bindingdb\_ic50} &
    \texttt{<protein>} \(+\) \texttt{<mol>} \(\rightarrow\) scalar &
    BindingDB \\
    & \texttt{bindingdb\_ec50} &
    \texttt{<protein>} \(+\) \texttt{<mol>} \(\rightarrow\) scalar &
    BindingDB \\
    & \texttt{bindingdb\_kd} &
    \texttt{<protein>} \(+\) \texttt{<mol>} \(\rightarrow\) scalar &
    BindingDB \\
    \midrule
    \multirow{1}{*}{cls} &
    \texttt{rna\_protein\_rpi} &
    \texttt{<rna>} \(+\) \texttt{<protein>} \(\rightarrow\) 0/1 &
    Biology-Instr. \\
    \bottomrule
  \end{tabularx}
  \endgroup
\end{table}

\subsection{Entity-level splitting and leakage control}
\label{sec:data-split}
Random record-level splits systematically overestimate performance on biological data, because near-duplicate sequences and scaffolds recur across the split boundary. For natively constructed datasets, protein clusters and molecular scaffolds are assigned at the split level rather than the individual-record level.
Exact duplicates are removed before split assignment.
We audit all splits for overlap under the specified protein-cluster and molecular-scaffold criteria.
For DAVIS and KIBA, we retain the official DeepDTA folds for comparability and cross-deduplicate BindingDB against those test pairs.

\subsection{Stage-2 Consolidation Mixture}
\label{sec:data-mixture}
The Stage-2 consolidation mixture combines all scientific modalities with scientific-text and general-purpose corpora. It includes scientific multiple-choice and open-ended QA in the style of MMLU/MMMU~\cite{mmlu_pro,mmmu_pro}, multimodal QA, general instruction-following, mathematics, code, agent and GUI-grounding data, general image--text and OCR data, and the scientific modality data described above. To prevent high-volume general corpora from dominating training, we down-sample large sources relative to scientific data while keeping smaller scientific benchmarks at full size.

\section{Training Strategy}
\label{sec:training}
\model\ follows a \textbf{two-stage modality-then-language curriculum}. Stage 1 aligns each scientific modality with the frozen Qwen3-VL backbone. For the sequence, molecular, and weather branches, the trainable components comprise the corresponding encoder, adapter, and applicable output decoder. MedSeg follows a different dual-path design: the Qwen3-VL vision-language pathway remains frozen, while the Qwen-to-SAM3 projection, all non-text SAM3 modules, and the auxiliary prediction head are optimized. This teaches each modality-specific component to translate its native signal into representations consumable by the language model while preserving the pretrained language prior. Stage~2 then loads and \textbf{freezes} the Stage-1 modality-specific components, \textbf{unfreezes} the LLM backbone, and mixes all scientific modalities with large-scale scientific-text and general instruction corpora. This consolidation stage improves scientific-text benchmark performance while retaining the multimodal capabilities acquired in Stage~1.

\begin{table}[t]
  \centering
  \caption{Optimization settings and trainable components for the two-stage training curriculum. Here, bs denotes per-device batch size and ga denotes gradient accumulation. Bioseq. + Mol. comprises the RNA, protein, DNA, and molecule branches.}
  \label{tab:training_config}
  \begingroup
  \footnotesize
  \setlength{\tabcolsep}{4pt}
  \renewcommand{\arraystretch}{1.15}
  \begin{tabularx}{\linewidth}{
    @{}
    >{\raggedright\arraybackslash}p{0.25\linewidth}
    >{\centering\arraybackslash}p{0.09\linewidth}
    >{\centering\arraybackslash}p{0.10\linewidth}
    >{\raggedright\arraybackslash}X
    @{}}
    \toprule
    \textbf{Modality / mixture} & \textbf{LR} &
    \textbf{bs \(\times\) ga} & \textbf{Trainable components} \\
    \midrule
    \multicolumn{4}{@{}l}{\textbf{Stage 1: Scientific-interface alignment}} \\
    \addlinespace[1pt]
    Bioseq. + Mol. &
    \(5\times10^{-5}\) & \(16\times1\) &
    Encoders + adapters; decoders as applicable \\
    Weather & \(1\times10^{-6}\) & \(1\times1\) &
    Encoder + adapter + decoder \\
    MedSeg & \(2\times10^{-6}\) & \(1\times2\) &
    SAM3 vision encoder + adapter + segmentation decoder \\
    \midrule
    \multicolumn{4}{@{}l}{\textbf{Stage 2: Shared-backbone consolidation}} \\
    \addlinespace[1pt]
    Mixed modalities & \(1\times10^{-6}\) & \(1\times4\) &
    Qwen3-VL LLM backbone + \texttt{lm\_head} \\
    \bottomrule
  \end{tabularx}
  \endgroup
\end{table}

\subsection{Stage 1: Independent Modality-component Training}
\label{sec:stage1}
In Stage~1, each active scientific modality is trained against the frozen Qwen3-VL language backbone. The trainable modules include the corresponding domain encoder, adapter, and, when applicable, modality decoder, as summarised in Table~\ref{tab:training_config}. RNA and protein are trained jointly in the Stage-1 run used for the reported mixed model because RNA--protein interaction data require both encoders to be instantiated and supervised in the same forward graph. DNA, molecule, weather, and medical segmentation modules are trained or loaded from their corresponding modality-specific Stage-1 checkpoints. Protein--molecule binding and enzyme datasets are used as mixed corpus in Stage~2 rather than in Stage-1.
Across Stage~1 runs, we use AdamW with weight decay $0.01$, a cosine learning-rate schedule with linear warm-up, gradient clipping at global norm $1.0$, \texttt{bf16} mixed precision, gradient checkpointing, and DeepSpeed ZeRO-2. The per-modality learning rates, batch sizes, and trainable components are listed in Table~\ref{tab:training_config}.

\paragraph{RNA and protein joint training.} RNA and protein are trained jointly because the RNA--protein interaction dataset contains paired RNA and protein inputs. This requires both modality encoders to be active in the same forward graph, while the frozen LLM provides the shared conditioning backbone. To stabilise distributed training with RNA-only, protein-only, and paired batches, the sampler keeps each accumulation window single-kind and synchronises the micro-batch type across data-parallel ranks.

\paragraph{Weather rollout training.} 
\label{sec:weathertrain} 
The weather component is trained with truncated autoregressive rollout. At each iteration, the model is unrolled for several consecutive $6$-hour steps by feeding its own predictions back as input, and gradients are back-propagated only through the final step. Lead-time conditioning is sampled across short forecast horizons, with lead-dependent loss scaling applied to the forecast loss. We also maintain an exponential moving average of the weather weights and apply patch-wise spatial dropout as augmentation.

\paragraph{Medical-segmentation training}
\label{sec:medsegtrain} 
MedSeg processes each image through two parallel visual pathways. The frozen Qwen3-VL vision-language pathway produces instruction-conditioned joint image-text hidden states, which are mapped by a trainable two-layer projection into the SAM3 text-embedding space. All non-text components of SAM3, including its vision backbone, segmentation decoder, and prediction heads, are jointly optimized with the Qwen-to-SAM3 projection, while the native SAM3 text encoder remains frozen. We further introduce a lightweight auxiliary meta-object classification head to provide coarse semantic supervision.

\subsection{Stage 2: Joint Seven-modality Consolidation}
\label{sec:stage2}
In Stage~2, all Stage~1 modality components, including encoders, adapters, and decoders, are loaded into a fresh base model and kept \emph{frozen}. We unfreeze only the Qwen3-VL LLM backbone and \texttt{lm\_head}, and train them with all seven modalities activated using the consolidation mixture described in Section~\ref{sec:data-mixture}. This stage aims to improve scientific-text and general reasoning performance while preserving the modality-specific skills acquired in Stage~1.

\paragraph{Optimisation.} As summarised in Table~\ref{tab:training_config}, Stage~2 freezes all Stage-1 modality components and optimises only the Qwen3-VL LLM backbone and \texttt{lm\_head}. We use AdamW with cosine scheduling, \texttt{bf16} mixed precision, gradient checkpointing, and DeepSpeed ZeRO-2. Because biological, weather, and MedSeg samples follow structurally different collators and loss paths, we use single-modality micro-batches and synchronise the micro-batch type across data-parallel ranks.  

% To stabilise ZeRO-2 training under these heterogeneous branches, we keep the backward graph topology consistent by attaching a zero-valued anchor to the shared trainable parameters and evaluating modality losses in a fixed order, with inactive branches multiplied by zero-valued weights rather than skipped. This preserves consistent gradient reduction order across ranks while allowing different modality groups to participate in the same accumulation window.

\subsection{Training objectives and supervision masks}
\label{sec:loss}
The training mixture contains heterogeneous objectives, including text generation, biological sequence modelling, molecular generation, weather forecasting, and medical-image segmentation. We optimise a unified objective with a text loss and one modality-specific loss for each active modality:
\begin{equation}
\mathcal{L}
  \;=\;
  \mathcal{L}_{\text{text}}
  \;+\;
  \sum_{m \in \mathcal{M}} s_m \,\mathcal{L}_{m},
  \qquad
  s_m \in \{0,1\},
\label{eq:totalloss}
\end{equation}
where $\mathcal{M}=\{\text{RNA},\text{mol},\text{weather},\text{MedSeg}\}$, and $s_m$ indicates whether modality $m$ is active in the current optimisation step. Active modality losses use equal outer weights, while structured losses such as weather forecasting and MedSeg retain the internal weights defined in their decoder sections.

The text loss $\mathcal{L}_{\text{text}}$ is the standard causal language-modelling cross-entropy over assistant answer tokens, with prompt tokens, modality placeholders, and domain-output positions masked using \texttt{ignore\_index}$=-100$. For modality understanding tasks, encoders and adapters are trained through the task-specific supervision signal, such as textual answer likelihood or dataset-specific classification/regression losses. In the reported configuration, sequence generation is supervised by two modality-specific heads: a bias-free RNA linear head over shared-backbone hidden states and an independent molecular autoregressive decoder over a decoupled SMILES vocabulary. Both heads read the LLM hidden states at their domain-output positions, apply a one-token causal shift, and compute cross-entropy against domain-native labels. DNA and protein decoding heads are supported by the framework but are not activated in the reported experiments.

\section{Experiment}
\label{sec:exp}

\subsection{Experimental Setup}

\subsubsection{Implementation Details}
\paragraph{Default checkpoint and optimization.}
Unless otherwise stated, all main experiments use the default Stage-2 unified checkpoint obtained after the two-stage training pipeline described in Section~\ref{sec:training}.
Optimization and distributed-training settings follow the configuration reported in Section~\ref{sec:training}.
The default evaluation setting uses a maximum context length of $32{,}768$ tokens.
Ablation studies, when reported, explicitly specify any deviation from this checkpoint, context length, or training configuration.

\paragraph{Hardware and distributed training.}
All training runs are conducted on NVIDIA H200 GPUs with 141\,GB of memory. We use PyTorch together with DeepSpeed ZeRO Stage~2, which shards optimizer states and gradients across data-parallel workers to reduce memory redundancy, and launch jobs with \texttt{torchrun} over multiple 8-GPU nodes. The language backbone uses FlashAttention-2 for efficient attention computation, while the native vision tower uses scaled-dot-product attention for numerical stability in the segmentation pathway. The Stage-1 single-domain runs are distributed over 48 to 248 GPUs depending on the memory and sequence-length requirements of each modality, and the Stage-2 seven-modality consolidation run is trained on 288 GPUs (36 nodes). For large multi-node jobs, we reduce the ZeRO communication bucket sizes to improve inter-node efficiency, and we disable communication-computation overlap to ensure stable numerical behaviour for medical segmentation.

\paragraph{Evaluation configuration.}
Table~\ref{tab:eval_config} summarises the decoding and context-length settings used for all reported benchmarks.
Values correspond to the CLI arguments actually passed by the invocation wrappers used to produce the reported numbers; per-task overrides are indicated in the same row.
All inference runs use \texttt{bf16} mixed precision, share the same fixed random seed, and share the same trained Stage-2 checkpoint. We use two inference paths depending on the task. Tasks that require the scientific modality encoders or decoders (DNA, RNA, protein, and molecule understanding and generation, as well as weather and segmentation) are run with the full modality-augmented model. Text-only and image-text benchmarks that exercise only the shared language backbone are served through vLLM with \texttt{gpu\_mem\_util=0.95} for throughput. Decoding hyperparameters for each group are listed in Table~\ref{tab:eval_config}.

\begin{table*}[t]
\centering
\scriptsize
\setlength{\tabcolsep}{5pt}
\renewcommand{\arraystretch}{1.15}
\caption{Evaluation configuration for \model\ across benchmark groups.
``$T$'' is sampling temperature, ``$p$'' and ``$k$'' are top-$p$ and top-$k$ truncation, ``pp'' and ``rp'' are presence and repetition penalties.}
\label{tab:eval_config}
\begin{tabular}{lll}
\toprule
\textbf{Group} & \textbf{Decoding} & \textbf{Max model len} \\
\midrule
\multicolumn{3}{l}{Scientific capability. Default: greedy (do\_sample=False).} \\
\midrule
Biology sequence group  & (shared defaults) &  32{,}768\\
Molecule & (shared defaults) & 32{,}768 \\
Cross-modal & (shared defaults) & 32{,}768 \\
Weather (ERA5) & AR rollout, base lead $\in\{6,12,18,24\}$\,h $\rightarrow 240$\,h & --- \\
Medical Seg & SAM3-head cls+box+mask (no LM decoding) & --- \\
\midrule
\multicolumn{3}{l}{General capability. Default: $T{=}1.0,\,p{=}1.0,\,k{=}40,\,\text{pp}{=}2.0,\,\text{rp}{=}1.0$.} \\
\midrule
Default group & (shared defaults) & 40{,}960 \\
IMO-AnsBench & (shared defaults) & 65{,}536 \\
RefCOCO-avg & greedy ($T{=}0$) & 16{,}384 \\
\bottomrule
\end{tabular}
\end{table*}

\subsubsection{Benchmarks}
\label{sec:benchmarks}
Our benchmark covers six scientific domains and scientific text. Unless otherwise stated, models are evaluated on held-out \texttt{test} splits, with task-specific evaluation settings summarized in Table~\ref{tab:eval_config}. Table~\ref{tab:benchmark_inventory_bio_mol} summarises the biological and molecular benchmarks used in our evaluation, covering DNA, RNA, protein, cross-modal, and molecule tasks across classification, regression, and generation settings.

\begin{table*}[t]
\centering
\scriptsize
\setlength{\tabcolsep}{5pt}
\renewcommand{\arraystretch}{1.12}
\caption{Benchmark inventory for biological and molecular evaluation tasks.}
\label{tab:benchmark_inventory_bio_mol}
\begin{tabular}{llp{5.2cm}l}
\toprule
\textbf{Domain} & \textbf{Task} & \textbf{Description} & \textbf{Task type} \\
\midrule
\multirow{6}{*}{DNA}
& EMP & Epigenetic Marks Prediction & Binary classification \\
& PD300 & Promoter Detection 300 & Binary classification \\
& CPD & Core Promoter Detection & Binary classification \\
& TB-H & TF Binding Sites Detection Human & Binary classification \\
& TB-M & TF Binding Sites Detection Mouse & Binary classification \\
& EA & Enhancer Activity Prediction & Regression \\
\midrule
\multirow{7}{*}{RNA}
& ncRNA & Non-coding RNA Function Classification & Multi-class classification \\
& APA & APA Isoform Prediction & Regression \\
& MRL & Mean Ribosome Loading Prediction & Regression \\
& PRS & Programmable RNA Switches & Regression \\
& Modif & Modification Prediction & Multi-class classification \\
& CRI-On & CRISPR On Target Prediction & Regression \\
& Toehold & Toehold Switch Design & Generation \\
\midrule
\multirow{5}{*}{Protein}
& Sta & Stability Prediction & Regression \\
& Flu & Fluorescence Prediction & Regression \\
& Ther & Thermostability Prediction & Regression \\
& EC & Enzyme Commission Number Prediction & Multi-label classification \\
& Sol & Solubility Prediction & Binary classification \\
\midrule
\multirow{3}{*}{Cross-modal}
& AAN & Antibody-Antigen Neutralization & Binary classification \\
& RPI & RNA-Protein Interaction Prediction & Binary classification \\
& EPI & Enhancer-Promoter Interaction Prediction & Binary classification \\
\midrule
\multirow{9}{*}{Molecule}
& ClinTox & ClinTox Classification & Binary classification \\
& SIDER & SIDER Classification & Binary classification \\
& BBBP & BBBP Classification & Binary classification \\
& HIV & HIV Classification & Binary classification \\
& ESOL & ESOL Regression & Regression \\
& Lipophilicity & Lipophilicity Regression & Regression \\
& ADMET & ADMET Classification & Binary classification \\
& SmolDesign & SMILES Design & Generation \\
& MoleHB & MoleHB Regression & Regression \\
\bottomrule
\end{tabular}
\end{table*}

\paragraph{RNA and DNA.}
Most RNA and DNA benchmarks are drawn from the Biology-Instructions suite~\cite{biology_instructions}. As summarised in Table~\ref{tab:benchmark_inventory_bio_mol}, they cover regulatory DNA tasks, RNA function and activity prediction, RNA--protein interaction, and RNA sequence generation across classification, regression, and generation settings. Inputs are raw nucleotide sequences or sequence pairs processed by the modality-specific DNA and RNA encoders described in Section~\ref{subsec:input_pathway}. Regression targets are evaluated on their raw physical scales, except for the binned enhancer-activity variant, which discretises labels into $100$ equal-frequency bins and maps predictions back to bin means. We note that the siRNA Efficiency Prediction task from the Biology-Instructions suite is excluded from Table~\ref{tab:benchmark_inventory_bio_mol}, as its underlying raw data is no longer available for download.

% ----------------------------------------------------------------------
\paragraph{Proteins.}
% ----------------------------------------------------------------------
Protein benchmarks are also drawn from the Biology-Instructions suite~\cite{biology_instructions}. As summarised in Table~\ref{tab:benchmark_inventory_bio_mol}, they cover protein-property regression, solubility classification, antibody--antigen neutralization, and multi-label enzyme-function annotation. Inputs are amino-acid sequences, with paired sequences used for antibody--antigen neutralization, and outputs include scalar targets, binary labels, or EC-code sets evaluated under the task-provided metrics.

% ----------------------------------------------------------------------
\paragraph{Molecules.}
% ----------------------------------------------------------------------
Molecule benchmarks cover ADMET classification, molecular property regression, MoleculeNet endpoints, and text-to-SMILES generation across classification, regression, and generation settings, as summarised in Table~\ref{tab:benchmark_inventory_bio_mol}. Prediction tasks use molecular graphs derived from SMILES strings, while the generation task maps text descriptions to SMILES. The SMolInstruct tasks are based on instruction-formatted MoleculeNet~\cite{moleculenet} and ChEBI-20/MolT5~\cite{molt5_chebi20} data~\cite{smolinstruct}, while ADMET and property-regression tasks follow TDC~\cite{tdc} and MoleHB~\cite{suiren}, respectively. 

% ----------------------------------------------------------------------
\paragraph{Earth science.}
% ----------------------------------------------------------------------
The Earth-science benchmark uses ERA5 global reanalysis data~\cite{era5}. Given one $0.25^\circ$ global weather snapshot, the task is to forecast future atmospheric and surface states. We train from 2002 January to 2023 June and evaluate on a temporally disjoint hold-out from 2023 July, with forecasts rolled out every $6$ hours up to $240$ hours.

\paragraph{Medical image segmentation.}
BiomedParse~\cite{biomedparse} is a text-grounded medical-image segmentation benchmark with image--prompt pairs and binary mask targets. We evaluate at $2016\times2016$ resolution on the pooled test set of $102{,}855$ pairs across nine imaging modalities: CT, MRI, X-ray, Ultrasound, Dermoscopy, Endoscopy, Pathology, Fundus, and OCT.

% ----------------------------------------------------------------------
\paragraph{Scientific text.}
% ----------------------------------------------------------------------
Scientific-text capability is evaluated on two public, evaluation-only benchmarks, \textit{MMLU-Pro}~\cite{mmlu_pro} and \textit{MMMU-Pro}~\cite{mmmu_pro}.
\textit{MMLU-Pro} is a $12{,}032$-question text multiple-choice benchmark across $14$ categories, with up to $10$ options per question, and is evaluated in a $5$-shot chain-of-thought setting.
\textit{MMMU-Pro} is a multimodal image--text multiple-choice benchmark; we evaluate its standard ($10$ options) and vision subsets, each with $1{,}730$ questions, in a zero-shot chain-of-thought setting.
In both cases, the task is to select a single answer option.

\subsubsection{Evaluation metrics}
\label{sec:metrics}
We report the primary metric used in each result table rather than listing all auxiliary evaluation statistics. For classification tasks, we report Matthews correlation coefficient (MCC), accuracy, or AUROC, depending on the benchmark convention. Regression tasks are evaluated with $R^2$ or Spearman's $\rho$, and the protein EC multi-label task is evaluated with $F_{\max}$. Native RNA and molecule generation are evaluated with exact match (EM), recovery-style sequence metrics, and fingerprint Tanimoto similarity (FTS), as appropriate to the output space. Dense-output tasks use domain-standard metrics: Earth-system forecasting is evaluated with latitude-weighted RMSE and anomaly correlation coefficient (ACC), while biomedical segmentation is evaluated with Dice score. General capability retention is reported using the official benchmark scores for each text, vision, reasoning, and coding evaluation.

\subsubsection{Competitors.}
\label{sec:competitors}
We compare \model\ against four families of baselines: general multimodal LLMs, scientific LLMs on biology understanding, specialist chemistry baselines, and modality-specific specialists.
These baselines are chosen to position \model\ along well-defined axes rather than to construct an exhaustive leaderboard.

\paragraph{General multimodal LLMs.}
For text and multimodal tasks, we compare \model\ with two general-purpose multimodal LLMs.
\emph{Qwen3-VL-8B-Instruct}~\cite{qwen3vl} is the direct backbone of \model\ (Section~\ref{sec:arch}).
This same-scale comparison tests whether integrating six scientific encoders and Stage-2 consolidation preserves the base VLM's general multimodal capability.
\emph{Intern-S1-Pro} is a trillion-parameter scientific mixture-of-experts model with an explicit ``thinking'' configuration for reasoning-heavy tasks.
Its roughly $100\times$ larger parameter budget provides a high-capacity reference for where a compact unified model can track a much larger reasoning-optimised system and where the gap remains.

\paragraph{Scientific LLMs on biology understanding.}
On the Biology-Instructions understanding suite (Table~\ref{tab:bio_understanding_combined}), we compare \model\ against two LLM-based references at markedly different capacity: \emph{Biology-Instructions}~\cite{biology_instructions} (the released ChatMultiOmics checkpoint, a Llama-3.1-8B-Instruct fine-tune that represents DNA, RNA, protein and multi-molecular sequences directly as text tokens without modality-specific encoders) and \emph{Intern-S1-Pro} (the trillion-parameter text-token-based scientific LLM reference introduced above).
The Biology-Instructions comparison isolates the benefit of \model's dedicated biological encoders at a matched $8$B parameter scale, while Intern-S1-Pro bounds how far a much larger text-token-only scientific LLM can go on the same tasks.
We do not report per-task specialist SOTAs on this suite, since \model's core comparison target here is other unified LLM-based models rather than task-tailored specialists trained end-to-end on a single benchmark.

\paragraph{Specialist chemistry baselines.}
For molecule benchmarks, we compare \model\ with standard chemistry-specific baselines for each suite.
These baselines are dedicated molecular models rather than general-purpose LLMs, and are selected to match the reference set commonly used by each benchmark family.

For \textbf{MoleHB}, we compare against \emph{Mole-BERT}, \emph{UniMol-v1}, and \emph{UniMol-v2}, together with the leaderboard SOTA, \emph{Suiren} in its conformer-averaged variant~\cite{suiren}.
Since Suiren is also the molecule encoder used by \model\ (Section~\ref{sec:arch}, Table~\ref{tab:modules}), this comparison effectively tests the effect of \model's shared instruction backbone and routing on top of a fixed molecular representation.

For \textbf{ADMET}, we compare against \emph{ChemProp}, a widely used D-MPNN framework for molecular property prediction, as well as \emph{DeepAuto-QSAR} and \emph{Uni-QSAR}, two QSAR-competition-tuned systems designed for ADMET-style leaderboards.
We also report the specialist SOTA on the same suite.
These baselines are all per-endpoint tuned, with task-specific heads or hyperparameter searches, and thus represent the strongest dedicated-model comparator for a unified instruction-driven model.

For \textbf{SMolInstruct}~\cite{smolinstruct} understanding tasks, we compare against \emph{Uni-Mol}~\cite{unimol_v1}, which is used by LlaSMol~\cite{smolinstruct} as the task-specific specialist for every MoleculeNet-style property-prediction endpoint~\cite{moleculenet}, together with \emph{LlaSMol} as an instruction-tuned chemistry LLM at a comparable parameter budget. For the SMolInstruct text-to-SMILES generation split, the task-specific specialist is \emph{MolT5}~\cite{molt5_chebi20}, again following LlaSMol's compared-model setup, we additionally include \emph{ChemLLM} in this generation comparison.

% \paragraph{Modality-specific baselines.}
\paragraph{Weather baselines.}
For weather forecasting, we compare \model\ against the operational \emph{ECMWF HRES}~\cite{hres} deterministic forecast, the physics-based numerical weather prediction (NWP) system used by the European Centre for Medium-Range Weather Forecasts, which is the standard baseline adopted by data-driven medium-range global forecast models such as Pangu-Weather~\cite{pangu-weather} and GraphCast~\cite{graphcast}.
This places \model\ against a production-grade NWP system rather than another data-driven model, and lets us report ML-vs-NWP gap and skill horizon on the same latitude-weighted metrics.

\paragraph{Medical segmentation baselines.} For medical segmentation, we compare \model\ with six segmentation networks: BiomedParse~\cite{biomedparse}, SAM~\cite{sam}, MedSAM~\cite{medsam}, SAM3~\cite{sam3}, DINO+SAM, and DINO+MedSAM~\cite{dinov2}. These baselines cover text-grounded biomedical segmentation, general and medical promptable segmentation, concept-driven segmentation, and DINOv2-conditioned mask decoding.

\subsection{Experimental Results}
\model\ is a unified scientific multimodal model designed to support both scientific understanding and generation across diverse scientific domains.
We therefore evaluate \model\ along three complementary axes that collectively probe the quality and generality of its shared multimodal representation.
The first axis focuses on scientific understanding tasks, where paired text and scientific-modality inputs, including DNA, RNA, proteins, and molecular graphs, are mapped to textual answers (Section~\ref{subsubsec:sci-understanding}).
The second axis evaluates scientific-modality generation, where text prompts are decoded into native-format scientific outputs through modality-specific decoder heads.
We compare \model\ with corresponding specialist generative models (Sections~\ref{subsubsec:sci-generation} and~\ref{subsubsec:end2end}).
The third axis focuses on text-only and image-text inputs with textual outputs, examining whether joint training across seven scientific and general modalities preserves the pretrained backbone's general knowledge, reasoning, grounding, coding, and OCR capabilities (Section~\ref{subsubsec:general}).

\subsubsection{Scientific Understanding Results}
\label{subsubsec:sci-understanding}

\paragraph{Biological Sequence Understanding.}
Table~\ref{tab:bio_understanding_combined} evaluates DNA, RNA, protein, and paired-sequence cross-modality understanding across 20 benchmarks, comparing \model\ against Biology-Instructions and Intern-S1-Pro.

Overall, \model\ performs strongly on classification-oriented biological tasks, especially those dominated by sequence-level or motif-like signals. In DNA, it obtains the best results among the compared LLM-based models on EMP, PD300, CPD, and TB-M, and is nearly tied with Intern-S1-Pro on TB-H. In RNA, it achieves the strongest results on non-coding RNA function classification and RNA modification prediction. These results indicate that the modality-specific encoders and adapters provide an effective component for biological sequence understanding, particularly when the target signal can be captured from local or medium-range sequence patterns.

Regression-oriented tasks show more variable behaviour. \model\ is competitive on several protein regression benchmarks, such as stability prediction and CRI-On prediction, but remains behind Intern-S1-Pro on thermostability and below the best compared model on several RNA regression tasks such as MRL and PRS. A similar pattern appears in DNA enhancer-activity prediction. Since these tasks require precise scalar prediction rather than categorical decision boundaries, they may benefit from dedicated numeric regression heads or stronger task-specific supervision, which are not the focus of the current configuration.

On paired-sequence cross-modality tasks, \model\ shows its strongest result on RNA--protein interaction prediction, reaching $76.49$ MCC and outperforming both Intern-S1-Pro and Biology-Instructions. It also substantially improves over Biology-Instructions on antibody--antigen neutralization, although Intern-S1-Pro remains slightly higher. In contrast, all compared LLM-based models perform near chance on enhancer--promoter interaction prediction, suggesting that this task remains difficult for sequence-only multimodal LLM pipelines and may require additional regulatory or structural priors.

Aggregating across all biological understanding benchmarks, \model\ achieves the best score on $9/20$ tasks and matches or exceeds Intern-S1-Pro on $10/20$, while outperforming Biology-Instructions on $16/20$ tasks. The last row of Table~\ref{tab:bio_understanding_combined} summarises this comparison as a single scalar: the ``average'' row reports the macro average over all 20 tasks with equal per-task weight. Under this metric, \model\ reaches $60.11$, outperforming Intern-S1-Pro by $8.14$ points ($51.97$) and the same-scale Biology-Instructions baseline by $22.67$ points ($37.44$), despite Intern-S1-Pro being approximately two orders of magnitude larger in total parameters. The main strength is concentrated in classification and interaction-prediction settings, whereas the remaining gaps are concentrated in scalar regression and structure-dependent pairwise tasks. These results motivate future extensions with dedicated regression heads and additional biological supervision for tasks requiring quantitative or higher-order structural reasoning.

\begin{table*}[t]
\centering
\scriptsize
\setlength{\tabcolsep}{5pt}
\renewcommand{\arraystretch}{1.12}
\caption{Performance comparison across biological sequence and cross-modality understanding tasks. \underline{Underlines} denote second-best results, and \textbf{bold} denotes best results.}
\label{tab:bio_understanding_combined}
\begin{tabular}{ll l ccc}
\toprule
\textbf{Domain} & \textbf{Task} & \textbf{Metric} &
\cellcolor{teal!8}\textbf{\model~(11B)} & \textbf{Biology-Instructions~(8B)} & \textbf{Intern-S1-Pro~(1T)} \\
\midrule
\multirow{6}{*}{DNA}
& EMP & MCC & \cellcolor{teal!8}\textbf{71.99} & 3.64 & \underline{14.02} \\
& PD300 & MCC & \cellcolor{teal!8}\textbf{91.17} & 58.18 & \underline{82.65} \\
& CPD & MCC & \cellcolor{teal!8}\textbf{66.35} & 44.54 & \underline{54.60} \\
& TB-H & MCC & \cellcolor{teal!8}\underline{54.01} & 24.45 & \textbf{54.11} \\
& TB-M & MCC & \cellcolor{teal!8}\textbf{65.91} & 39.91 & \underline{60.80} \\
& EA & PCC & \cellcolor{teal!8}52.64 & \underline{53.28} & \textbf{55.16} \\
\midrule
\multirow{6}{*}{RNA}
& ncRNA & Acc & \cellcolor{teal!8}\textbf{91.46} & \underline{63.09} & 34.50 \\
& APA & $R^2$ & \cellcolor{teal!8}\underline{79.87} & 59.01 & \textbf{82.95} \\
& MRL & $R^2$ & \cellcolor{teal!8}35.54 & \underline{47.64} & \textbf{52.41} \\
& PRS & $R^2$ & \cellcolor{teal!8}25.99 & \underline{26.57} & \textbf{33.97} \\
& Modif & AUC & \cellcolor{teal!8}\textbf{96.03} & \underline{59.06} & 57.77 \\
& CRI-On & Spearman's $\rho$ & \cellcolor{teal!8}\textbf{28.76} & -0.02 & \underline{15.69} \\
\midrule
\multirow{5}{*}{Protein}
& Sta & Spearman's $\rho$ & \cellcolor{teal!8}\textbf{70.63} & 60.25 & \underline{60.82} \\
& Flu & Spearman's $\rho$ & \cellcolor{teal!8}\underline{70.12} & 2.57 & \textbf{78.14} \\
& Ther & Spearman's $\rho$ & \cellcolor{teal!8}\underline{46.37} & 45.07 & \textbf{59.56} \\
& EC & Fmax & \cellcolor{teal!8}\underline{68.65} & 19.79 & \textbf{72.70} \\
& Sol & Acc & \cellcolor{teal!8}\underline{67.26} & 63.02 & \textbf{67.60} \\
\midrule
\multirow{3}{*}{Cross-modal}
& AAN & MCC & \cellcolor{teal!8}\underline{42.96} & 1.06 & \textbf{44.76} \\
& RPI & MCC & \cellcolor{teal!8}\textbf{76.49} & \underline{74.26} & 58.51 \\
& EPI & MCC & \cellcolor{teal!8}\underline{-0.03} & \textbf{3.37} & -1.30 \\
\midrule
\multicolumn{2}{c}{\textbf{average}} & & \cellcolor{teal!8}\textbf{60.11} & 37.44 & \underline{51.97} \\
\bottomrule
\end{tabular}%
\end{table*}

\paragraph{Molecular Understanding.}
Table~\ref{tab:molecule_property_tasks} and Table~\ref{tab:smolinstruct_understanding_tasks} report \model's performance across three complementary chemistry benchmark families: MoleHB, ADMET, and SMolInstruct.

On SMolInstruct, \model\ is competitive across both classification and regression endpoints, achieving the best or tied-best result on $4/6$ tasks. It obtains the strongest results on BBBP, SIDER, and ESOL, ties Uni-Mol on HIV, and remains close on ClinTox and Lipophilicity. These results indicate that the molecule encoder and shared language backbone can support instruction-conditioned molecular property prediction across diverse endpoints.

On the broader MoleHB and ADMET suites, \model\ remains below the strongest chemistry-specialist models. It reaches a normalized average score of $0.8408$ on MoleHB, above UniMol-v2 and Mole-BERT but below Suiren, and obtains an ADMET average of $0.7304$, below purpose-built QSAR and ADMET systems. This gap reflects the breadth--specialization trade-off of a unified scientific model: \model\ covers molecular tasks together with other scientific modalities, whereas the strongest chemistry baselines are optimized specifically for molecular property prediction.

\begin{table*}[t]
\centering
\scriptsize
\setlength{\tabcolsep}{5pt}
\renewcommand{\arraystretch}{1.15}
\caption{Performance comparison on MoleHB and ADMET molecular understanding benchmarks. For MoleHB, we report normalized average scores over all sub-tasks due to the large number of fine-grained tasks. For ADMET, we report the direct average over all sub-tasks. \underline{Underlines} denote second-best results, and \textbf{bold} denotes best results.}
\label{tab:molecule_property_tasks}
\begin{tabular}{p{1.8cm}p{2.2cm}p{1.8cm}p{1.8cm}p{1.8cm}p{1.8cm}p{1.8cm}}
\toprule
\textbf{Dataset} & \textbf{Metric} & \cellcolor{teal!8}\textbf{\model} & \textbf{Suiren} & \textbf{Mole-BERT} & \textbf{UniMol-v1} & \textbf{UniMol-v2} \\
\midrule
MoleHB & Normalized Avg. & \cellcolor{teal!8}0.8408 & \textbf{0.9693} & 0.0597 & \underline{0.8980} & 0.7139 \\
\bottomrule
\end{tabular}
\begin{tabular}{p{1.8cm}p{2.2cm}p{1.8cm}p{1.8cm}p{1.8cm}p{1.8cm}p{1.8cm}}
\toprule
\textbf{Dataset} & \textbf{Metric} & \cellcolor{teal!8}\textbf{\model} & \textbf{Suiren} & \textbf{DeepAuto-QSAR} & \textbf{Uni-QSAR} & \textbf{ChemProp} \\
\midrule
ADMET & Avg. & \cellcolor{teal!8}0.7304 & \underline{0.8046}  & 0.7858 & \textbf{0.8084} & 0.7624\\
\bottomrule
\end{tabular}
\end{table*}

\begin{table}[t]
\centering
\scriptsize
\setlength{\tabcolsep}{5pt}
\renewcommand{\arraystretch}{1.15}
\caption{Performance comparison on SMolInstruct molecular understanding tasks. \underline{Underlines} denote second-best results, and \textbf{bold} denotes best results.}
\label{tab:smolinstruct_understanding_tasks}
\begin{tabular}{lcccc}
\toprule
\textbf{Task} & \textbf{Metric} & \cellcolor{teal!8}\textbf{\model} & \textbf{Uni-Mol} & \textbf{LlaSMol} \\
\midrule
BBBP classification & \multirow{4}{*}{Acc} & \cellcolor{teal!8}\textbf{96.95} & \underline{85.30} & 74.60 \\
ClinTox classification & & \cellcolor{teal!8}92.36 & \underline{92.40} & \textbf{93.10} \\
HIV classification & & \cellcolor{teal!8}\textbf{97.00} & \textbf{97.00} & \underline{96.70} \\
SIDER classification & & \cellcolor{teal!8}\textbf{71.00} & 70.00 & \underline{70.70} \\
\midrule
ESOL regression & \multirow{2}{*}{RMSE$\downarrow$} & \cellcolor{teal!8}\textbf{0.550} & \underline{0.819} & 1.150 \\
Lipophilicity regression & & \cellcolor{teal!8}\underline{0.628} & \textbf{0.612} & 1.010 \\
\bottomrule
\end{tabular}
\end{table}

\subsubsection{Scientific Generation Results}
\label{subsubsec:sci-generation}
Beyond understanding, \model\ supports native-format scientific generation through modality-specific decoders (Section~\ref{subsec:output-generation}). In this report, we evaluate two such capabilities: RNA sequence generation and molecule generation.

\paragraph{RNA Generation.}
Within the biology suite, we evaluate native RNA generation on an internal toehold-switch design split, where a trigger sequence and linker are mapped to a full switch RNA sequence by the RNA decoder. Since no matched generalist or specialist baselines are available, we report standalone performance only. \model\ achieves near-saturated scores, with BLEU $99.996$ and per-position recovery $99.998\%$, consistent with the largely deterministic structure of this design task, in which the switch stem is close to a reverse-complement expansion of the trigger combined with a fixed linker/scaffold. These results indicate that the shared-backbone RNA decoder can recover the target switch sequence with near single-nucleotide precision on this split.

\paragraph{Molecule Generation.}
We evaluate native molecular generation on SMolInstruct SMILES design tasks. It achieves a validity of $89.09\%$, an exact-match (EM) rate of $22.22\%$, and a fingerprint Tanimoto similarity (FTS) of $61.96$, compared with $95.3/31.7/73.2$ for MolT5~\cite{molt5_chebi20} and $99.7/19.2/61.7$ for LlaSMol. Although \model\ has lower syntactic validity than LlaSMol, it achieves higher EM and FTS. The remaining gap to MolT5 is most visible on exact reconstruction and fingerprint similarity, suggesting that future improvements should focus on both increasing SMILES validity and better constraining generation toward the target molecular structure.

\begin{table}[t]
\centering
\scriptsize
\setlength{\tabcolsep}{6pt}
\renewcommand{\arraystretch}{1.15}
\caption{Performance comparison on SMolInstruct molecule generation. We report validity, exact match (EM), and fingerprint Tanimoto similarity (FTS). \underline{Underlines} denote second best results, \textbf{bold} denotes best results.}
\label{tab:molecule_generation_tasks}
\begin{tabular}{lcccc}
\toprule
\textbf{Metric} & \cellcolor{teal!8}\textbf{\model} & \textbf{MolT5} & \textbf{LlaSMol} & \textbf{ChemLLM} \\
\midrule
Validity & \cellcolor{teal!8}89.09 & \underline{95.3} & \textbf{99.7} & 4.3 \\
EM & \cellcolor{teal!8}\underline{22.22} & \textbf{31.7} & 19.2 & 0.9 \\
FTS & \cellcolor{teal!8}\underline{61.96} & \textbf{73.2} & 61.7 & 14.3 \\
\bottomrule
\end{tabular}
\end{table}

\paragraph{Cross-modal molecule generation.}
The enzyme-catalysis data introduced in Section~\ref{sec:data-crossmodal} provides a direct test of cross-modal native generation, where protein context is used to condition molecular outputs. We evaluate this capability on two enzyme-centred tasks: cofactor prediction from enzyme sequence, and substrate-to-product prediction conditioned on enzyme sequence, substrate molecules, and a curated Rhea reaction identifier. Since these internally constructed tasks have no matched generalist or specialist baselines, we report representative cases rather than model-versus-model scores, treating predictions as correct when they match the curated database answers under RDKit-canonical SMILES or InChI.
Figure~\ref{fig:molecular_generation_cases} shows one example per task produced by the released checkpoint. In the cofactor case the model recovers glutathione, the tripeptide cofactor annotated for the enzyme in UniProt, reproducing the reference structure exactly. In the substrate-to-product case the model hydrolyses the ester bond of a monoacylglycerol substrate and returns free oleate and glycerol, which is exactly the transformation catalysed by the lipase. These examples illustrate that the molecule decoder can generate molecular outputs conditioned on protein context within the shared multimodal framework, using the same unified checkpoint without task-specific tuning.

\begin{figure*}[t]
\centering

\begin{tcolorbox}[enhanced, colback=gray!4, colframe=gray!55!black, boxrule=0.4pt,
  left=6pt, right=6pt, top=4pt, bottom=4pt, width=0.98\textwidth]
\footnotesize
\textbf{(a) Cofactor prediction.}

\smallskip
\textbf{System.} You are an enzyme cofactor annotation expert. Given an enzyme protein sequence, output the complete set of UniProt-annotated cofactors for this sequence as SMILES. Cofactors may include metal ions; they are not limited to organic coenzymes. Output only SMILES separated by \texttt{'.'}, sorted lexicographically.

\smallskip
\textbf{Input.}\\
\texttt{<protein>}\\
Predict all UniProt-annotated cofactors required by this enzyme sequence, including metal ions. Output only the SMILES set.

\smallskip
{\itshape where} \texttt{<protein>} {\itshape is a 230-residue enzyme (MSTNSDLRVTLYTYFRSS\dots).}

\smallskip
\textbf{Model output.}\\
{\ttfamily [NH3+][C@@H](CCC(=O)N[C@@H](CS)C(=O)NCC(=O)[O-])C(=O)[O-]}

\smallskip
{\itshape Glutathione. Matches the UniProt-annotated cofactor exactly (RDKit-canonical identical).}
\end{tcolorbox}

\vspace{0.6em}

\begin{tcolorbox}[enhanced, colback=gray!4, colframe=gray!55!black, boxrule=0.4pt,
  left=6pt, right=6pt, top=4pt, bottom=4pt, width=0.98\textwidth]
\footnotesize
\textbf{(b) Substrate-to-product prediction.}

\smallskip
\textbf{System.} You are an enzymatic reaction annotation expert. Given an enzyme protein sequence, substrate molecule(s), and a Rhea reaction ID from a curated UniProt/Rhea annotation, predict the product molecule(s) for that specific reaction. Output only the product SMILES; multiple species are separated by \texttt{'.'}.

\smallskip
\textbf{Input.}\\
\texttt{<protein>}\\
\texttt{<mol>}\\
This enzyme is annotated with Rhea:38488. For this specific Rhea reaction, predict the product molecule(s) from the substrate set. Output only the product SMILES.

\smallskip
{\itshape where} \texttt{<protein>} {\itshape is a 1450-residue lipase (MESWPGVSLVGLLLLLL\dots) and} \texttt{<mol>} {\itshape is the substrate set} {\ttfamily CCCCCCCC/C=C\textbackslash CCCCCCCC(=O)OCC(O)CO.O} {\itshape (monoacylglycerol + water).}

\smallskip
\textbf{Model output.}\\
{\ttfamily CCCCCCCC/C=C\textbackslash CCCCCCCC(=O)[O-].OCC(O)CO.[H+]}

\smallskip
{\itshape Oleate + glycerol. Matches the curated Rhea:38488 product exactly.}
\end{tcolorbox}

\caption{Representative native molecular generation cases. 
(a) Enzyme sequence to cofactor SMILES. 
(b) Enzyme and substrate molecules to product SMILES.}
\label{fig:molecular_generation_cases}
\end{figure*}

\subsubsection{Dense Scientific Output Generation}
\label{subsubsec:end2end}

\begin{figure}[!t]
\vskip 0.2in
\begin{center}
\centerline{\includegraphics[width=\columnwidth]{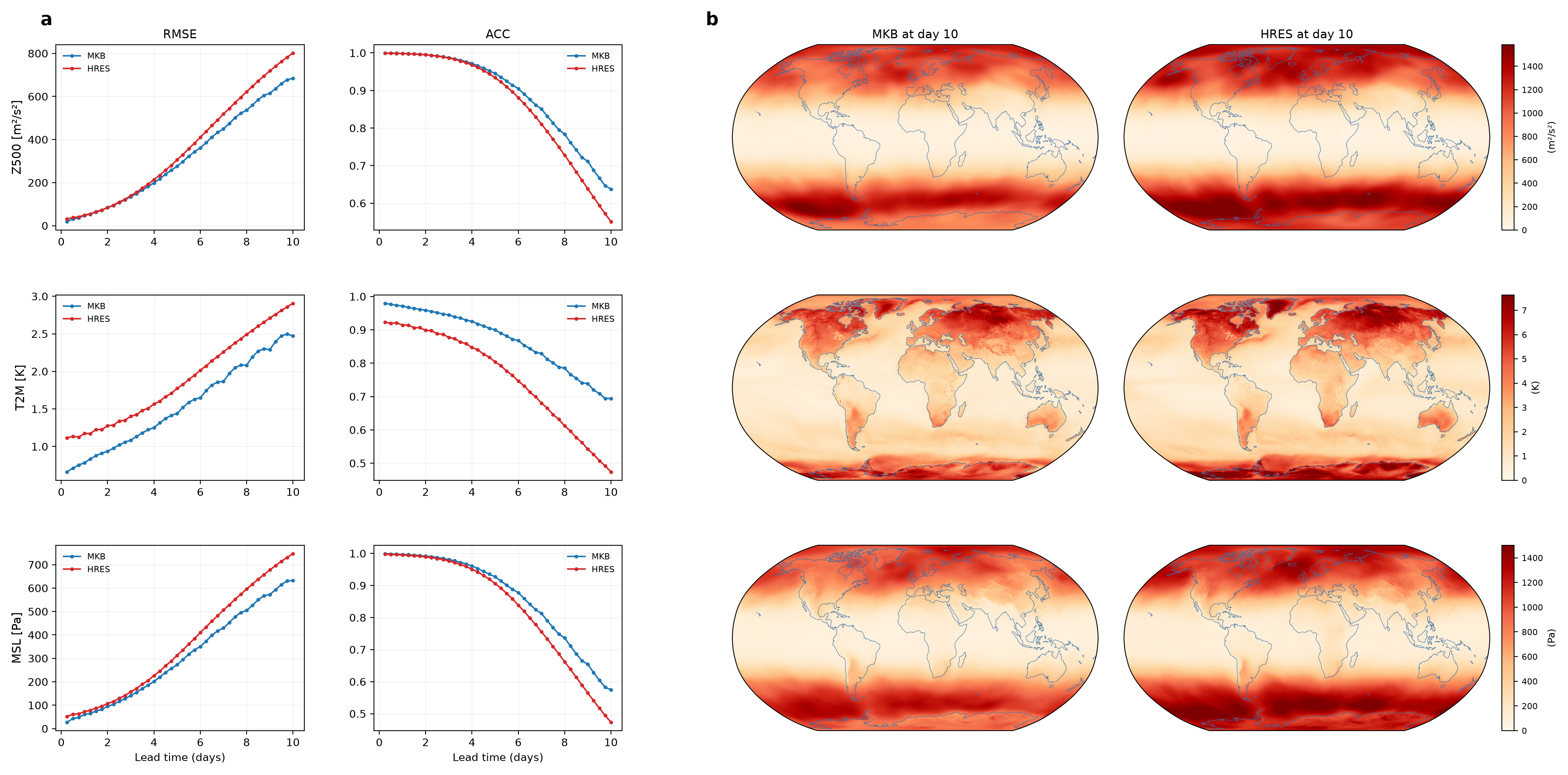}}
\caption{\model\ vs.\ ECMWF HRES on global ERA5 forecasting. \textbf{(a)} Latitude-weighted RMSE and anomaly correlation coefficient (ACC) as a function of lead time (up to $10$ days) for $500$-hPa geopotential (Z500), 2-metre temperature (T2M), and mean sea-level pressure (MSL). \textbf{(b)} Absolute-error maps at day $10$ for the same three variables.}
\label{fig:weather}
\end{center}
\vspace{-0.8cm}
\end{figure}

\paragraph{Earth-Science Forecasting.}
Figure~\ref{fig:weather} reports medium-range global forecasting results on the ERA5 hold-out set at $0.25^{\circ}$ resolution. Forecasts are initialised on the 2023 July to 2024 June hold-out at $0000$/$1200$ UTC and rolled out autoregressively every $6$ hours up to $240$\,h. We evaluate against ERA5 using latitude-weighted RMSE and anomaly correlation coefficient (ACC), with ACC computed relative to the ERA5 1990--2019 climatology. Following common practice in data-driven medium-range forecasting~\cite{pangu-weather,graphcast,fourcastnet}, we report three representative variables covering upper-air dynamics, near-surface impact, and large-scale pressure: $500$-hPa geopotential (Z500), 2-metre temperature (T2M), and mean sea-level pressure (MSL).

Figure~\ref{fig:weather}(a) shows that \model\ generally outperforms HRES across Z500, T2M, and MSL, with larger gains at longer lead times. For Z500, the two systems are similar through the early forecast range, but \model\ maintains lower error and higher ACC after roughly day $4$. By day $10$, it reaches about $680~\mathrm{m^2/s^2}$ RMSE and $0.64$ ACC, compared with about $800~\mathrm{m^2/s^2}$ and $0.55$ for HRES. Similar long-range improvements appear for MSL, where \model\ reaches about $625~\mathrm{Pa}$ RMSE at day $10$ versus about $740~\mathrm{Pa}$ for HRES, while T2M shows the most persistent advantage across the full $10$-day horizon, reaching day-$10$ RMSE of about $2.5~\mathrm{K}$ versus $2.9~\mathrm{K}$ for HRES. Across these variables, the slower medium-range degradation of RMSE and ACC indicates stronger forecast skill from \model\ at longer lead times.

The day-$10$ absolute-error maps in Figure~\ref{fig:weather}(b) show a consistent spatial pattern. HRES has larger errors across mid-latitude and polar regions for all three variables, and especially over northern-hemisphere land areas for T2M. \model's errors are lower in magnitude, with residual hotspots for Z500 concentrated over the Southern Ocean storm-track band and the high-latitude polar caps, and, for T2M, over northern-hemisphere land in a spatially reduced version of the HRES pattern.

\paragraph{Medical-Image Segmentation.}

\begin{table*}[t]
\centering
\scriptsize
\setlength{\tabcolsep}{4pt}
\renewcommand{\arraystretch}{1.15}
\caption{Dice score comparison across medical imaging modalities. We report the average Dice score (\%) for each method. The best result in each row is highlighted in \textbf{bold}, and the second-best result is \underline{underlined}.}
\label{tab:medical_segmentation_results}
\begin{tabular}{lccccccc}
\toprule
\textbf{Modality} & \cellcolor{teal!8}\textbf{\model} & \textbf{BiomedParse} & \textbf{DINO+MedSAM} & \textbf{DINO+SAM} & \textbf{MedSAM} & \textbf{SAM3} & \textbf{SAM} \\
\midrule
All & \cellcolor{teal!8}\textbf{91.20} & \underline{90.73} & 15.37 & 15.10 & 83.55 & 35.40 & 71.29 \\
CT & \cellcolor{teal!8}\textbf{93.36} & \underline{92.25} & 9.59 & 10.34 & 83.87 & 28.93 & 74.10 \\
MRI & \cellcolor{teal!8}\textbf{85.29} & \underline{85.25} & 13.28 & 12.39 & 75.90 & 53.64 & 68.34 \\
OCT & \cellcolor{teal!8}\underline{85.31} & \textbf{86.63} & 6.68 & 6.98 & 56.26 & 8.69 & 55.99 \\
X-ray & \cellcolor{teal!8}\underline{98.02} & \textbf{98.28} & 37.22 & 30.63 & 97.75 & 39.96 & 81.35 \\
Dermoscopy & \cellcolor{teal!8}\textbf{98.08} & 97.11 & 81.28 & 78.29 & \underline{97.35} & 51.47 & 88.23 \\
Endoscopy & \cellcolor{teal!8}\textbf{97.39} & 96.77 & 25.01 & 24.54 & \underline{97.05} & 38.82 & 92.88 \\
Fundus & \cellcolor{teal!8}\underline{91.33} & \textbf{91.50} & 3.19 & 2.73 & 88.06 & 18.58 & 57.16 \\
Pathology & \cellcolor{teal!8}\textbf{87.29} & \underline{81.57} & 25.38 & 24.69 & 43.44 & 26.08 & 42.06 \\
Ultrasound & \cellcolor{teal!8}\underline{90.54} & \textbf{91.03} & 17.12 & 22.91 & 89.76 & 5.23 & 57.47 \\
\bottomrule
\end{tabular}
\end{table*}

Table~\ref{tab:medical_segmentation_results} reports mean Dice on the BiomedParse \texttt{test} splits, aggregated over $102{,}855$ image--prompt pairs spanning nine imaging modalities. On the pooled \emph{All} split, \model\ achieves the best average Dice score ($91.20$), slightly ahead of BiomedParse ($90.73$) and substantially above MedSAM, SAM, SAM3, and the DINO-conditioned variants. Across modalities, \model\ obtains the best score on $5/9$ subsets (CT, MRI, Dermoscopy, Endoscopy, and Pathology) and ranks second on the remaining four (OCT, X-ray, Fundus, and Ultrasound), with only small gaps to the leading method on those subsets. The largest margin appears on Pathology ($87.29$ vs.\ $81.57$, $+5.72$ Dice), followed by CT ($93.36$ vs.\ $92.25$, $+1.11$ Dice). 

The comparison with SAM3 is particularly informative because \model\ uses SAM3-style dense visual features but adds instruction-conditioned semantic guidance through the shared multimodal backbone. The large gap between off-the-shelf SAM3 ($35.40$) and \model\ ($91.20$) highlights the importance of semantic conditioning for text-prompted biomedical segmentation. The low scores of the DINO-conditioned variants further suggest that generic visual features alone are insufficient for this setting.

Overall, these results show that the dual-path biomedical segmentation pathway can match or exceed dedicated segmentation specialists while remaining part of the same unified checkpoint used for the other scientific modalities.

\begin{figure*}[!t]
\centering
\includegraphics[width=\textwidth]{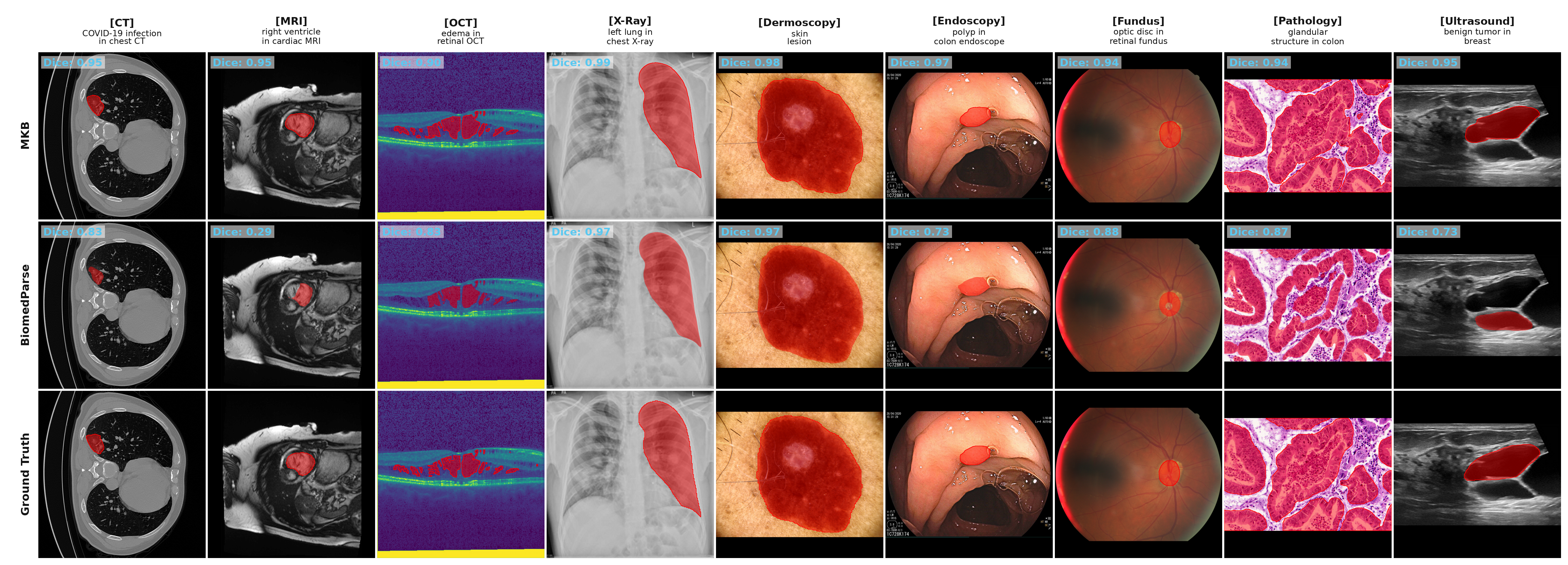}
\caption{Qualitative text-prompted segmentation comparison between \model\ and BiomedParse across the nine medical-imaging modalities of Table~\ref{tab:medical_segmentation_results}.
%For each modality we display one case and the accompanying per-instance Dice scores are indicative rather than modality averages.
Row~1: \model's predictions, Row~2: BiomedParse predictions, Row~3: ground-truth masks; per-instance Dice is overlaid on each non-GT tile.
Targets, left to right, are: \emph{COVID-19 infection} (chest CT, COVID-19-CT), \emph{right ventricle} (cardiac MRI, ACDC), \emph{retinal edema} (OCT, OCT-CME), \emph{left lung} (chest X-ray, COVID-QU-Ex), \emph{skin lesion} (dermoscopy, ISIC), \emph{colon polyp} (endoscopy, NeoPolyp), \emph{optic disc} (retinal fundus, G1020), \emph{glandular structure} (H\&E pathology, GlaS), and \emph{benign tumor} (breast ultrasound, BreastUS).}
\label{fig:medseg_qualitative}
\end{figure*}

\begin{table*}[t]
\centering
\scriptsize
\setlength{\tabcolsep}{8pt}
\renewcommand{\arraystretch}{1.15}
\caption{General capability comparison between \model\ and its shared Transformer backbone, Qwen3-VL-8B-Ins. \textbf{Bold} denotes the better result, with ties bolded for both models.}
\label{tab:general_capability}
\begin{tabular}{llcc}
\toprule
\textbf{Dataset} & \textbf{Metric}
& \cellcolor{teal!8}\textbf{\model}
& \textbf{Qwen3-VL-8B-Ins} \\
\midrule
MMLU-Pro & \multirow{4}{*}{Accuracy} & \cellcolor{teal!8}73.31\% & \textbf{73.36\%} \\

MMMU-Pro & & \cellcolor{teal!8}\textbf{57.60\%} & 57.29\% \\

AIME-2025 & & \cellcolor{teal!8}\textbf{46.67\%} & 43.33\% \\

ScreenSpot V2
&
& \cellcolor{teal!8}\textbf{92.30\%}
& \textbf{92.30\%} \\
\midrule

IMO-Answer-Bench
& Accuracy (avg@8)
& \cellcolor{teal!8}\textbf{34.94\%}
& 34.63\% \\

RefCOCO-avg
& Acc@IoU$\geq$0.5
& \cellcolor{teal!8}88.00\%
& \textbf{88.01\%} \\

IFBench
& Strict prompt-level accuracy
& \cellcolor{teal!8}\textbf{32.33\%}
& \textbf{32.33\%} \\
\midrule

OCRBench V2 ENG
& \multirow{2}{*}{OCRBench V2 score}
& \cellcolor{teal!8}57.40\%
& \textbf{57.50\%} \\

OCRBench V2 CHN
&
& \cellcolor{teal!8}\textbf{63.90\%}
& 63.80\% \\
\midrule

SArena (Icon)
& SArena-Icon score
& \cellcolor{teal!8}71.49\%
& \textbf{74.83\%} \\

LCB V6
& pass@1
& \cellcolor{teal!8}\textbf{50.43\%}
& 50.33\% \\
\bottomrule
\end{tabular}
\end{table*}

Figure~\ref{fig:medseg_qualitative} compares \model's text-prompted predictions with BiomedParse against the ground-truth masks on one case per modality drawn from the pooled BiomedParse test split. Reading each column top-to-bottom, we compare how closely the top-row \model\ mask and the middle-row BiomedParse mask reproduce the bottom-row annotation in shape, extent, and location.

Reading the columns from left to right, the differences against the ground truth are most visible on four modalities. On \textit{[CT]}, \model\ tightly outlines the annotated COVID-19 infection while BiomedParse bleeds into neighbouring parenchyma; on \textit{[MRI]}, \model\ recovers the right-ventricle blood pool while BiomedParse marks the surrounding myocardial wall instead; on \textit{[Pathology]}, \model\ traces the glandular structures faithfully while BiomedParse fragments them; on \textit{[Ultrasound]}, \model\ hits the correct lesion while BiomedParse mislocalises the tumour to an unrelated region of the frame. On the remaining modalities (\textit{[OCT]}, \textit{[X-Ray]}, \textit{[Dermoscopy]}, \textit{[Endoscopy]}, \textit{[Fundus]}), both models produce masks that closely follow the annotation, with only minor boundary differences between the two.

Overall, the qualitative comparison supports the quantitative trend of Table~\ref{tab:medical_segmentation_results}: \model\ matches or exceeds BiomedParse across the nine BiomedParse modalities, with the largest visible gains concentrated on targets whose extent requires instruction-conditioned semantic reasoning rather than local texture cues alone.

\subsubsection{General Capability Results}
\label{subsubsec:general}
Table~\ref{tab:general_capability} compares \model\ with its shared Transformer backbone, Qwen3-VL-8B-Instruct, on general capability benchmarks. Overall, \model\ largely preserves the backbone's general abilities after scientific multimodal training. Across the 11 evaluated tasks, \model\ matches or surpasses the backbone on 7 tasks and is slightly lower on the remaining 4. Most differences are marginal and within seed- or sampling-level variation. The only notable drop is observed on SArena-Icon, where \model\ scores 71.49\% compared with 74.83\% for Qwen3-VL-8B-Instruct, possibly due to limited SVG-oriented supervision in the current Stage-2 mixture. These results indicate that \model\ retains the general-purpose capabilities of its 8B VLM backbone while adding scientific multimodal understanding and generation abilities.

\section{Conclusion}
\label{sec:conclusion}
Existing scientific AI systems generally lack a unified capability to understand and generate heterogeneous scientific modalities, with most models remaining specialised to individual domains or limited to text-centric interactions. To address this gap, we introduce \model, a unified scientific multimodal model that supports cross-modal understanding, reasoning, and modality-native generation across six scientific domains within a shared modelling framework.

\model\ combines modality-tailored encoders, adapters, and decoders with a shared autoregressive backbone, allowing heterogeneous scientific inputs to be represented in structurally appropriate forms and jointly modelled within a shared representation space. The results demonstrate the feasibility of this shared-backbone paradigm, showing that a unified scientific model can remain competitive across diverse scientific tasks while largely preserving general-purpose capabilities.

Despite these promising results, performance remains less consistent on tasks requiring precise scalar prediction, particularly ADMET-related endpoints. This limitation suggests that unified scientific multimodal modelling remains an evolving direction, especially for tasks that require highly accurate quantitative prediction.

\section*{Author Contributions}

\vspace{-0.5em}
\noindent\rule{\textwidth}{0.4pt}
\vspace{0.4em}
{\small
\noindent\textbf{Core Contributors}\\
Hesen Chen, Xinyu Su, Xiaomeng Yang, Yuetan Lin

\vspace{0.75em}
\noindent\textbf{Data Contributors}\\
{Protein data:} Zixiong Yang\\
{Molecule data:} Junyi An, Fenglei Cao\\
{Scientific text:} Yifeng Jiao\\
{RNA data:} Yunqi Zhang, Yuan Cheng

\vspace{0.75em}
\noindent\textbf{Core Contributor Leads and Corresponding Authors}\\
Zhiyu Tan, Hao Li

\vspace{0.75em}
\noindent\textbf{Executive Sponsors}\\
Libo Wu, Yuan Qi
}

\vspace{0.4em}
\noindent\rule{\textwidth}{0.4pt}

\bibliographystyle{plainnat}
\bibliography{references}

\end{document}